\pdfoutput=1
\documentclass[sigconf, nonacm]{acmart}

\setcopyright{acmlicensed}
\copyrightyear{2027}
\acmYear{2027}
\acmDOI{XXXXXXX.XXXXXXX}
\acmConference[WWW '27]{The ACM Web Conference 2027}{2027}{TBA}
\acmISBN{978-1-4503-XXXX-X/2027/XX}

\usepackage{booktabs}
\usepackage{nicefrac}
\usepackage{colortbl}
\usepackage{multirow}
\usepackage{dsfont}
\usepackage{pifont}
\usepackage{makecell}
\usepackage{listings}
\usepackage{tcolorbox}
\usepackage[normalem]{ulem}
\useunder{\uline}{\ul}{}
\usepackage{algorithm}
\usepackage{algpseudocode}
\usepackage{subcaption}
\usepackage{enumitem}
\usepackage{xurl}
\definecolor{ivory2}{RGB}{212,212,212}

\tcbset{
    promptstyle/.style={
        colback=gray!5,
        colframe=gray!60,
        colbacktitle=black!80,
        fonttitle=\bfseries,
        boxrule=0.5pt,
        arc=2pt,
        left=4pt,
        right=4pt,
        top=4pt,
        bottom=4pt
    }
}

\begin{document}

\title{TrajDLM: Topology-Aware Block Diffusion Language Model for Trajectory Generation}

\author{Wilson Wongso\texorpdfstring{\textsuperscript{1,$\ast$}}{}, Lihuan Li\texorpdfstring{\textsuperscript{1}}{}, Arian Prabowo\texorpdfstring{\textsuperscript{1}}{}, Xiachong Lin\texorpdfstring{\textsuperscript{1}}{}, Baiyu Chen\texorpdfstring{\textsuperscript{1}}{}, Hao Xue\texorpdfstring{\textsuperscript{1,2}}{}, Flora Salim\texorpdfstring{\textsuperscript{1}}{}}
\affiliation{%
  \institution{\texorpdfstring{\textsuperscript{1}}{}University of New South Wales, Sydney, Australia}
  \institution{\texorpdfstring{\textsuperscript{2}}{}Hong Kong University of Science and Technology (Guangzhou), Guangzhou, China}
  \country{}}

\renewcommand{\shortauthors}{Wongso et al.}
\makeatletter
\def\@authorfont{\large\sffamily}
\g@addto@macro\@authornotes{\stepcounter{footnote}\footnotetext{Corresponding author: \href{mailto:w.wongso@unsw.edu.au}{w.wongso@unsw.edu.au}}}
\makeatother
\hypersetup{pdfauthor={Wilson Wongso, Lihuan Li, Arian Prabowo, Xiachong Lin, Baiyu Chen, Hao Xue, Flora Salim}}

\begin{abstract}
    Generating high-fidelity synthetic Web of Things (WoT) telemetry, such as GPS trajectories, is increasingly important for transportation, urban planning, and location-aware services, particularly when privacy concerns restrict access to real-world mobility data. Existing trajectory generation methods face a trade-off between efficiency and road network fidelity: continuous-space methods generate trajectories efficiently but ignore road topology, while topology-aware methods rely on search-based autoregressive decoding that limits generation speed. We propose \textbf{TrajDLM}, a topology-aware trajectory generation framework based on block diffusion language models. TrajDLM represents trajectories as sequences of discrete road segments, combining a block diffusion backbone, graph-based road network representations, and topology-constrained sampling to generate realistic trajectories with topologically valid road transitions. To address the tendency of supervised models to favor near-shortest routes, we introduce \textbf{TrajRFT}, a rejection-sampling fine-tuning stage that improves route fidelity by training on self-generated routes selected for their similarity to real trajectories. Across three city-scale vehicle telemetry datasets, TrajDLM closely matches real-world mobility in population-level distributions, local route similarity, and route diversity, while generating trajectories up to $7.5\times$ faster than the strongest baseline. It also demonstrates zero-shot transfer to unseen transport modes. These results establish block diffusion with topology-constrained sampling and rejection-sampling fine-tuning as an efficient approach to generating realistic synthetic mobility telemetry for location-aware Web services and WoT applications.
\end{abstract}

\maketitle

\section{Introduction}

Modeling human mobility supports applications such as traffic forecasting~\cite{lv2014traffic,jin2023spatio} and urban planning~\cite{zheng2023spatial,zheng2023road,jiang2016route}. GPS trajectories collected by mobile devices and location-aware applications provide valuable telemetry for Web-connected systems within the Web of Things (WoT). However, privacy concerns increasingly restrict access to real-world mobility data~\cite{chen2024trajectory,yuan2025breaking}, limiting the development of location-aware services. High-fidelity \emph{synthetic} trajectory generation offers a promising alternative~\cite{zhu2023difftraj,cao2025hoser}, enabling service development and evaluation without exposing real users' movement traces. To serve as a reliable substitute, synthetic trajectories must preserve both population-level mobility distributions and fine-grained route patterns shaped by the underlying road network.

Prior work has explored a wide range of deep generative models for trajectory generation, including Generative Adversarial Networks~\cite{yu2017seqgan,liu2018trajgans,feng2020learning,zhang2023dp,jiang2023continuous}, Variational Autoencoders~\cite{huang2019variational,10.1109/TKDE.2023.3312209}, diffusion models~\cite{zhu2023difftraj}, flow-matching models~\cite{li2026trajflow}, and autoregressive transformers~\cite{wang2024spatiotemporal,deng2025marionette,cao2025hoser}. However, these approaches largely fall into two categories. Efficient continuous-space generators such as DiffTraj~\cite{zhu2023difftraj} and TrajFlow~\cite{li2026trajflow} do not explicitly model the road network, producing trajectories that are not topology-aware. In contrast, topology-aware models such as TS-TrajGen~\cite{jiang2023continuous} and HOSER~\cite{cao2025hoser} rely on an iterative classical search algorithm at decoding time, which significantly limits generation speed. In practice, downstream applications such as city-scale simulation and what-if planning often require generating millions of high-fidelity trajectories under diverse conditions.

To address this gap, we propose \textbf{TrajDLM}, a topology-aware trajectory generation framework based on block diffusion language models. The core idea is to model trajectories as sequences of discrete road segments and generate them via iterative denoising, enabling high-fidelity synthesis while preserving road network structure. TrajDLM is motivated by three considerations: \ding{182}~Diffusion-based generation provides a framework for modeling stochastic mobility patterns through iterative refinement~\cite{yang2023diffusion,zhu2023difftraj}. \ding{183}~Modeling trajectories as discrete road segments treats road segments as the natural units of the road network~\cite{10.1145/3394486.3403043,10.1145/3347146.3359094,cao2025hoser}. \ding{184}~Discrete diffusion via language models generates many trajectories in parallel as a single batch, improving efficiency over search-based decoding~\cite{sahoo2024simple,nie2025large}.

However, naively applying discrete (masked) diffusion language models to trajectory generation presents several challenges. First, road segments are treated as independent tokens without explicit connectivity, making it difficult to preserve valid transitions and coherent routes. Second, the iterative denoising process is unconstrained by the road network, which can lead to invalid or fragmented trajectories. Finally, modeling road segments as standard text token embeddings neglects road network topology, limiting the model's ability to capture structural relationships between segments. Moreover, maximum-likelihood training favors shortest routes over the diverse paths taken by real drivers~\cite{cao2025hoser}, with early route deviations compounding into larger errors on long trips.

Building on these considerations, TrajDLM integrates four components: \ding{182}~\textbf{Block diffusion language model backbone}, modeling trajectories in block-wise units of road segments to preserve locality and coherence while enabling efficient generation; \ding{183}~\textbf{Topology-constrained sampling}, enforcing valid transitions from origin to destination; \ding{184}~\textbf{Road network encoder}, injecting topology-aware representations in place of standard token embeddings; and \ding{185}~\textbf{Trajectory rejection-sampling fine-tuning}, fine-tuning on self-generated routes selected for similarity to real trajectories. These components enable high-fidelity and efficient trajectory generation. Our contributions are as follows:

\begin{itemize}[leftmargin=*,nosep]
    \item \textbf{Topology-aware block diffusion for trajectory generation.} We introduce TrajDLM, the first block diffusion language model for trajectory generation, combining block-wise discrete diffusion with road segment-based trajectory modeling.
    \item \textbf{Graph-constrained trajectory synthesis.} We integrate graph-based road-network representations and propose topology-constrained sampling to enforce coherent trajectories.
    \item \textbf{Trajectory rejection-sampling fine-tuning.} We propose TrajRFT, which fine-tunes the model on high-quality self-generated routes to improve route diversity beyond shortest-path behavior.
    \item \textbf{High-fidelity and efficient generation.} TrajDLM closely matches real-world mobility across three city-scale vehicle telemetry datasets from the Web in terms of population-level distributions, local trajectory similarity, and route diversity, while being up to $7.5\times$ faster than HOSER and remaining competitive on GeoLife in a zero-shot setting.
\end{itemize}

\section{Related Work}

\paragraph{Trajectory Generation}

Deep generative models for trajectory generation have evolved from GAN-based approaches~\cite{yu2017seqgan,liu2018trajgans,feng2020learning,zhang2023dp,jiang2023continuous}, to autoencoders~\cite{huang2019variational,10.1109/TKDE.2023.3312209}, transformers~\cite{cao2021generating,wang2024spatiotemporal,deng2025marionette}, diffusion models~\cite{zhu2023difftraj}, and flow-matching methods~\cite{li2026trajflow}. These models differ not only in their generative paradigms but also in how trajectories are spatially represented. Earlier approaches model trajectories directly in continuous space~\cite{zhu2023difftraj,li2026trajflow,zhuunitraj}, generating sequences of GPS points. While these methods enable efficient generation, they do not explicitly model road network topology, allowing generated coordinates to deviate from valid roads or produce physically implausible routes~\cite{zhu2024controltraj}. To improve spatial grounding, more recent approaches discretize trajectories into spatial tokens using grid cells or hierarchical spatial cells~\cite{cao2021generating,chang2023contrastive,li2024t,li2025hit}. While discretization constrains trajectories to fixed spatial regions, these cells remain spatially arbitrary and do not naturally capture road network topology or semantics. Recent works \cite{jiang2023continuous,cao2025hoser} instead represent trajectories as sequences of road segments and employ graph neural networks (GNNs) to encode road network structure, capturing connectivity, geometry, and functional relationships across the road graph~\cite{kipf2016semi,brody2022how,10.1145/3347146.3359094,10.1145/3394486.3403043}. Building on these topology-aware trajectory representations, TrajDLM models trajectories as sequences of road segments and uses graph-based representations to keep them topologically coherent.

\paragraph{Discrete Diffusion}

Diffusion models have achieved remarkable success across a wide range of generative tasks and domains~\cite{sohl2015deep,rombach2022high,yang2023diffusion,kong2020diffwave,qin2023diffusion}, including continuous space trajectory generation~\cite{zhu2023difftraj,zhu2024controltraj}. While existing diffusion-based trajectory generation has thus far focused on continuous space, recent advances in diffusion language models~\cite{lou2023discrete,sahoo2024simple,li2025survey,nie2025large,ye2025dream,bie2025llada2,bie2026llada2} demonstrate that iterative denoising over discrete tokens can match autoregressive transformers in generation quality while enabling parallel sampling. Block diffusion language models (BDLMs)~\cite{arriola2025block} interpolate between autoregressive and diffusion language models through semi-autoregressive block-wise generation, combining long-range dependency modeling with efficient block-wise decoding, properties that align naturally with the requirements of trajectory generation. Motivated by this, TrajDLM is the first to apply BDLMs to trajectory generation under topological constraints.

\paragraph{Fine-Tuning on Self-Generated Samples}

Rejection-sampling fine-tuning (RFT)~\cite{yuan2023scalingrelationshiplearningmathematical} and reward-ranked fine-tuning (RAFT)~\cite{dong2023raft,xiong2025minimalist} improve a supervised model by generating multiple candidates per input, ranking them, and fine-tuning on selected outputs. Related self-training methods, such as ReST~\cite{gulcehre2023reinforced} and STaR~\cite{zelikman2022star}, also learn from model-generated samples but differ in their selection and training procedures. Unlike policy-gradient methods such as PPO~\cite{schulman2017proximal} and GRPO~\cite{shao2024deepseekmath}, these approaches optimize a supervised likelihood objective on filtered samples, without requiring a learned critic or explicit advantage estimation. TrajRFT adapts this approach to trajectory generation by sampling topology-constrained candidate routes and selecting those most similar to real trajectories.

\section{Preliminaries}
\label{sec:prelims}

\paragraph{Definition 1: Road Network}
We represent a road network as a directed graph $\mathcal{G} = \langle \mathcal{V}, \mathcal{E} \rangle$, where $\mathcal{V}$ denotes road segments and $\mathcal{E}$ represents intersections between adjacent road segments.

\paragraph{Definition 2: Trajectory}
A trajectory is defined as a sequence of road segments $\tau = [r_1, r_2, \dots, r_L]$, where $r_i \in \mathcal{V}$. Each transition satisfies $(r_{i-1}, r_i) \in \mathcal{E}$ for all $i \in \{2, \dots, L\}$.

\paragraph{Problem Statement: Conditional Trajectory Generation}
Given a dataset of trajectories $\mathcal{T}$, we aim to learn a generative model $G_\theta$ over road segment sequences conditioned on $(r_{\text{org}}, t_{\text{org}}, r_{\text{dest}})$, i.e., the origin segment, departure time, and destination segment, and trip attributes such as total distance $d_{\text{trip}}$, average segment distance $\bar{d}_{\text{seg}}$, trip duration $t_{\text{trip}}$, and average speed $v_{\text{avg}}$~\cite{zhu2023difftraj}. The model generates $\hat{\tau}$ such that $\hat{r}_1 = r_{\text{org}}$ and $\hat{r}_L = r_{\text{dest}}$.

\section{Methodology}

\subsection{Block Discrete Denoising Diffusion Language Model}
\label{sec:block-diffusion}

We represent trajectories as sequences of discrete road segments and model them using Block Discrete Denoising Diffusion Language Models (BD3-LM)~\cite{arriola2025block}, which combine autoregressive generation over blocks with discrete diffusion within each block.

\paragraph{Discrete Diffusion Process}

We adopt the Discrete Denoising Diffusion Probabilistic Model (D3PM) framework~\cite{austin2021structured} as instantiated in BD3-LM~\cite{arriola2025block}. Let $\tau_0 = [r_1, \dots, r_L]$ denote a clean trajectory over a vocabulary of size $|\mathcal{V}|$, augmented with a mask token $\texttt{[M]}$, and let $\tau_t = [\tau_t^1, \dots, \tau_t^L]$ be the latent at diffusion step $t \in \{0, \ldots, T\}$, where $\tau_0$ is the clean sequence and $\tau_T$ is fully corrupted. The forward process independently corrupts each token by replacing it with $\texttt{[M]}$; for consecutive steps $s$ and $t = s+1$, the per-position transition is:
\begin{equation}
    q(\tau_t^i \mid \tau_s^i) =
    \begin{cases}
        1           & \text{if } \tau_s^i = \tau_t^i = \texttt{[M]},                    \\
        1 - \beta_t & \text{if } \tau_t^i = \tau_s^i \neq \texttt{[M]},                 \\
        \beta_t     & \text{if } \tau_t^i = \texttt{[M]},\; \tau_s^i \neq \texttt{[M]}, \\
        0           & \text{otherwise},
    \end{cases}
\end{equation}
where $\beta_t \in (0,1)$ is the step-dependent masking rate, and the full-sequence transition factorizes as $q(\tau_t \mid \tau_s) = \prod_{i=1}^L q(\tau_t^i \mid \tau_s^i)$. A trained diffusion model $G_\theta$ reverses this process by learning to denoise $\tau_t$ back to $\tau_0$. Following D3PM, the reverse step factorizes independently across positions as $p_\theta(\tau_s \mid \tau_t) = \prod_{i=1}^L p_\theta(\tau_s^i \mid \tau_t)$, with each factor marginalizing over the predicted clean token $\tilde{\tau}_0^i$ via the tractable forward posterior:
\begin{equation}
    p_\theta(\tau_s^i \mid \tau_t) =
    \sum_{\tilde{\tau}_0^i}
    q(\tau_s^i \mid \tau_t^i, \tilde{\tau}_0^i)\,
    p_\theta(\tilde{\tau}_0^i \mid \tau_t),
\end{equation}
where $p_\theta(\tilde{\tau}_0^i \mid \tau_t)$ denotes the model's prediction of the clean $i$-th token given the noisy input.

\paragraph{Trajectory Blocks}

We partition a trajectory $\tau$ into $B$ contiguous blocks of length $L'$, with $B = L / L'$ and the $b$-th block defined as $\tau^b := [r_{(b-1)L'+1}, \dots, r_{bL'}]$. The log-likelihood factorizes autoregressively over blocks as $\log p_\theta(\tau) = \sum_{b=1}^B \log p_\theta(\tau^b \mid \tau^{<b})$, where $\tau^{<b}$ denotes all preceding blocks. Each block is denoised via the same per-position posterior as above, conditioned additionally on $\tau^{<b}$, giving $p_\theta(\tau_s^b \mid \tau_t^b, \tau^{<b})$.

\paragraph{Negative Evidence Lower Bound}

We train the model by minimizing the standard D3PM negative evidence lower bound (NELBO)~\cite{sohl2015deep, arriola2025block}. For a block $\tau^b$ conditioned on $\tau^{<b}$, the objective is:
\begin{equation}
    \begin{split}
        &\mathcal{L}(\tau^b \mid \tau^{<b}; \theta)
        =
        \mathbb{E}_q \Big[
            -\log p_\theta(\tau_0^b \mid \tau_1^b, \tau^{<b}) \\
        &\quad + \sum_{t=2}^{T}
            \mathrm{KL}\big(
            q(\tau_{t-1}^b \mid \tau_t^b, \tau_0^b)
            \,\|\,
            p_\theta(\tau_{t-1}^b \mid \tau_t^b, \tau^{<b})
            \big) \\
        &\quad + \mathrm{KL}\big(q(\tau_T^b \mid \tau_0^b) \,\|\, p(\tau_T^b)\big)
            \Big],
    \end{split}
\end{equation}
which decomposes across blocks as:
\begin{equation}
    \label{eq:bdnelbo}
    -\log p_\theta(\tau)
    \le
    \mathcal{L}_{\mathrm{BD}}(\tau; \theta)
    :=
    \sum_{b=1}^B \mathcal{L}(\tau^b \mid \tau^{<b}; \theta).
\end{equation}

\paragraph{Model Architecture}

The reverse process $p_\theta(\tau^b_s \mid \tau^b_t, \tau^{<b})$ is parameterized by a block diffusion language model (BDLM) $G_\theta$. We initialize $G_\theta$ from a pretrained BDLM. All road segment tokens $r_i \in \mathcal{V}$ are added to the tokenizer vocabulary, and the text embedding layer is resized accordingly. To incorporate road network topology, we replace the input embeddings of road tokens with road embeddings generated by the Road Network Encoder (Section~\ref{sec:rne}).

\subsection{Road Network Encoder}
\label{sec:rne}

To incorporate road network topology and semantics into the language model, we adapt the Road Network Encoder (RNE) from HOSER~\cite{cao2025hoser}, but retain only the road-level semantic representation without the zone-level component. We validate the contribution of the RNE in the ablation study (Section~\ref{sec:ablation}). Following HOSER's formulation, the road segments in $\mathcal{V}$ are treated as nodes in a graph, and the intersections between adjacent roads are treated as edges. The RNE encodes road segments and intersections separately, then fuses them via a graph attention network.

\paragraph{Road Segment Embedding}

For each road segment $r_i$, we construct its representation by combining a road segment ID with four attributes: length, highway type, longitude, and latitude. These components are independently embedded and concatenated to form the final road segment embedding $\boldsymbol{v}_i = \boldsymbol{v}_{\mathrm{ID}} \,\Vert\, \boldsymbol{v}_{\mathrm{len}} \,\Vert\, \boldsymbol{v}_{\mathrm{type}} \,\Vert\, \boldsymbol{v}_{\mathrm{lon}} \,\Vert\, \boldsymbol{v}_{\mathrm{lat}} \in \mathbb{R}^d$, where $\boldsymbol{v}_{(\cdot)}$ denotes the embedding of each feature and $\Vert$ indicates concatenation. Road IDs and highway types are encoded via learnable embeddings, while continuous attributes (length, longitude, latitude) are normalized and projected through linear layers.

\paragraph{Intersection Embedding}

To model directional connectivity, we define an edge-level representation for each ordered pair $(i, j)$ as $\boldsymbol{e}_{ij} = \mathds{1}_{ij} \,\Vert\, \boldsymbol{\phi}_{ij}$, where $\mathds{1}_{ij}$ denotes the reachability embedding (indicating whether $r_j$ is directly reachable from $r_i$ in $\mathcal{E}$), and $\boldsymbol{\phi}_{ij}$ denotes the steering-angle embedding derived from the normalized angular difference between the bearing angles of $r_i$ and $r_j$.

\paragraph{Graph Attention Network}

We employ GATv2~\cite{brody2022how} to fuse the road segment and intersection embeddings and propagate information over the road graph, enabling each segment representation to be contextually refined through its neighbors. At layer $(\ell+1)$, the segment representation is:
\begin{equation}
    \boldsymbol{v}_i^{(\ell+1)} = \sum_{j \in \mathcal{N}(i) \cup \{i\}} \alpha_{ij}^{(\ell)} \, \boldsymbol{v}_j^{(\ell)} \boldsymbol{\Theta}_{\mathrm{src}}^{(\ell)},
\end{equation}
where $\mathcal{N}(i)$ denotes the adjacent road segments of $r_i$. The attention coefficients are computed as:
\begin{equation}
    \alpha_{ij}^{(\ell)} = \mathrm{Softmax}\!\left( \sigma\!\left( \boldsymbol{v}_i^{(\ell)} \boldsymbol{\Theta}_{\mathrm{tgt}}^{(\ell)} + \boldsymbol{v}_j^{(\ell)} \boldsymbol{\Theta}_{\mathrm{src}}^{(\ell)} + \boldsymbol{e}_{ij} \right) \big(\boldsymbol{a}^{(\ell)}\big)^{\!\top} \right),
\end{equation}
where $\sigma$ is LeakyReLU, and $\boldsymbol{\Theta}_{\mathrm{src}}^{(\ell)}, \boldsymbol{\Theta}_{\mathrm{tgt}}^{(\ell)} \in \mathbb{R}^{d \times d}$ and $\boldsymbol{a}^{(\ell)} \in \mathbb{R}^d$ are learnable parameters. Multiple GATv2 layers are stacked to enrich topology representations $\boldsymbol{v}_i$ for every road segment $r_i$.

\paragraph{Road Token Embedding Injection}

To condition the BDLM on road network topology, we project the road representations $\boldsymbol{v}_i$ into the BDLM's input embedding space via a two-layer MLP, $\boldsymbol{z}_i = \mathrm{GELU}(\boldsymbol{v}_i \boldsymbol{W}_1 + \boldsymbol{b}_1) \boldsymbol{W}_2 + \boldsymbol{b}_2$, where $\boldsymbol{W}_1 \in \mathbb{R}^{d \times d_p}$, $\boldsymbol{W}_2 \in \mathbb{R}^{d_p \times d_\mathrm{LLM}}$, and $d_p$ is the projection hidden dimension. During the forward pass, the token embedding for each road segment token $r_i$ is replaced with $\boldsymbol{z}_i$, while embeddings for all non-road tokens remain unchanged.

\begin{figure*}
    \centering
    \includegraphics[width=\linewidth]{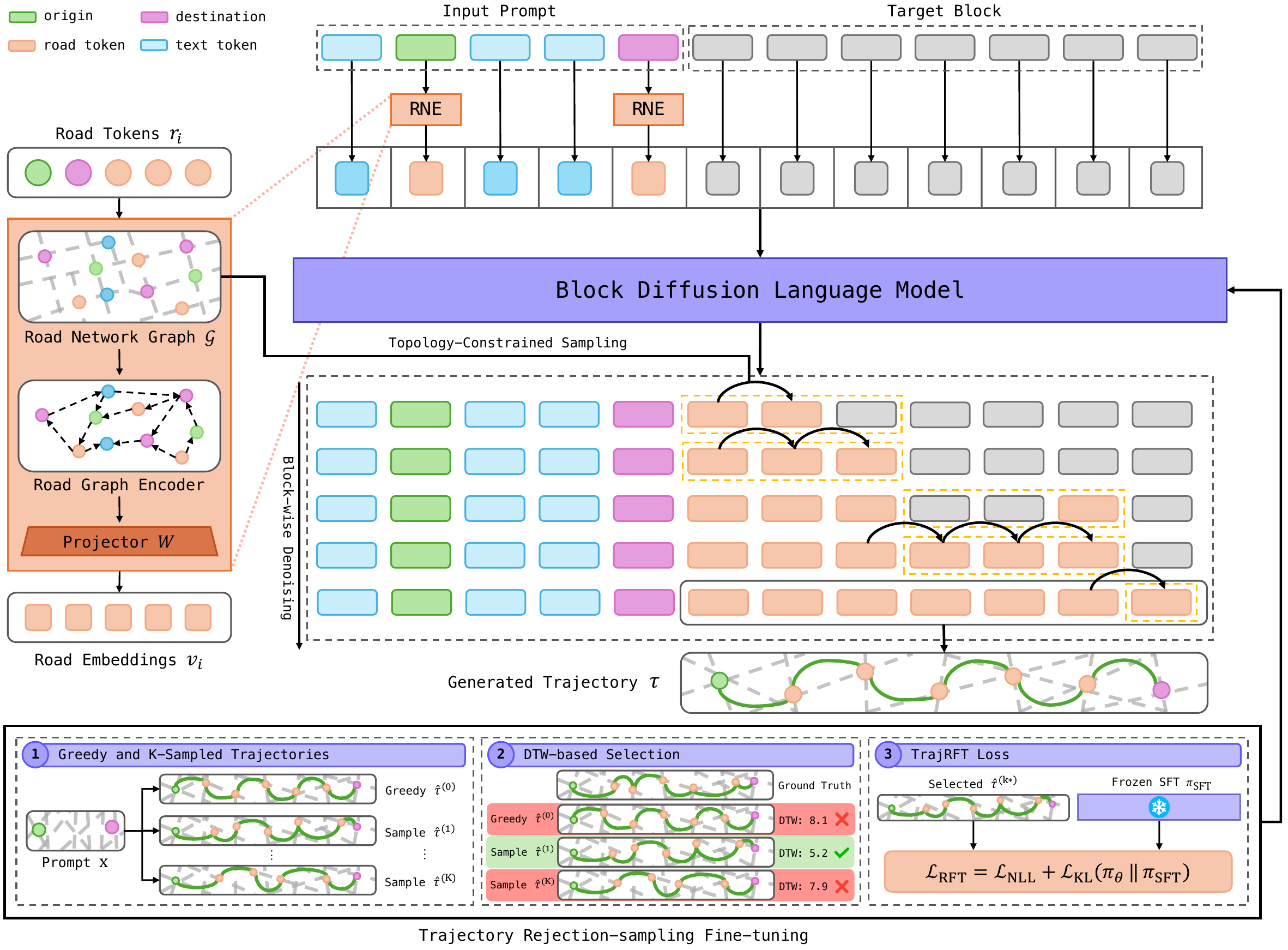}
    \caption{
        \textbf{Proposed TrajDLM architecture.} TrajDLM consists of a Block Diffusion Language Model backbone, a Road Network Encoder for topology-aware road embeddings, and a Topology-Constrained Sampling strategy for valid trajectory generation via block-wise denoising, trained in two stages: supervised fine-tuning followed by trajectory rejection-sampling fine-tuning.
    }
    \Description{Architecture diagram of TrajDLM. On the left, road tokens are encoded by a road network encoder: a road graph encoder over the road network graph followed by a projector produces road embeddings. At the top, an input prompt of text tokens with origin and destination road tokens is followed by a target block of mask tokens; road tokens in the prompt are replaced by their road network encoder embeddings before entering the block diffusion language model. Below the model, rows show block-wise denoising steps from t equals T to t equals 0, in which mask tokens are progressively replaced by road tokens that are each connected to the previous road segment under topology-constrained sampling. At the bottom, the generated trajectory is drawn as a connected path on the road network from the origin to the destination.}
    \label{fig:trajdlm-overview}
\end{figure*}

\subsection{Topology-Constrained Sampling}
\label{sec:tcd}

The NELBO objective trains the BDLM to model the empirical distribution of valid trajectories, but the reverse diffusion process itself does not enforce explicit topological constraints. A single incorrect road segment prediction may violate road connectivity, since consecutive segments $(\hat{r}_i, \hat{r}_{i+1})$ are not guaranteed to satisfy $(\hat{r}_i, \hat{r}_{i+1}) \in \mathcal{E}$. In addition, block-wise diffusion generates a fixed-length sequence, whereas real trajectories terminate naturally upon reaching the destination $r_{\text{dest}}$. To address these limitations, we introduce a topology-constrained sampling (TCS) strategy that augments the reverse process with \ding{182} adjacency-aware token restriction and left-to-right commitment, \ding{183} origin anchoring, \ding{184} reaching the destination, \ding{185} soft destination guidance, and \ding{186} revisit avoidance. Constraints \ding{182}--\ding{184} are hard: every generated trajectory starts at $r_{\text{org}}$, follows only adjacent road segments, and ends at $r_{\text{dest}}$ by construction. Constraints \ding{185}--\ding{186} only re-weight the valid candidates. The details and the pseudocode of topology-constrained sampling are provided in Appendix~\ref{app:tcs-pseudocode}.

\begin{table*}[tbp]
    \centering
    \scriptsize
    \caption{\textbf{Trajectory generation performance on Beijing, Porto, and San Francisco}, evaluated on the test split of each city. Models are grouped by whether they use a road network graph. TrajDLM is our full model with RNE, trained with SFT followed by TrajRFT. \textbf{Bold} marks the best score per column; {\ul underline} marks the second-best. $^\dagger$TrajSynVAE does not follow the road network, so nearly all of its road-segment $4$-grams are distinct and its $|\Delta$\textit{Div.}$|$ exceeds $2{,}000\%$.}
    \label{tab:main-results}
    \setlength{\tabcolsep}{1.45pt}
    \resizebox{\textwidth}{!}{%
    \begin{tabular}{lcc|cccc|cc|cccc|cc|cccc}
        \toprule
                          & \multicolumn{6}{c|}{\textbf{Beijing}} & \multicolumn{6}{c|}{\textbf{Porto}} & \multicolumn{6}{c}{\textbf{San Francisco}} \\ \cmidrule{2-19}
                          & \multicolumn{2}{c|}{\textbf{Global} ($\downarrow$)} & \multicolumn{4}{c|}{\textbf{Local} ($\downarrow$)} & \multicolumn{2}{c|}{\textbf{Global} ($\downarrow$)} & \multicolumn{4}{c|}{\textbf{Local} ($\downarrow$)} & \multicolumn{2}{c|}{\textbf{Global} ($\downarrow$)} & \multicolumn{4}{c}{\textbf{Local} ($\downarrow$)} \\ \cmidrule{2-19}
        \textbf{Method}   & \textbf{Dis.} & \multicolumn{1}{c|}{\textbf{Rad.}} & \textbf{Hau.} & \textbf{DTW} & \textbf{EDR} & \multicolumn{1}{c|}{\textbf{$|\Delta$Div.$|$}} & \textbf{Dis.} & \multicolumn{1}{c|}{\textbf{Rad.}} & \textbf{Hau.} & \textbf{DTW} & \textbf{EDR} & \multicolumn{1}{c|}{\textbf{$|\Delta$Div.$|$}} & \textbf{Dis.} & \multicolumn{1}{c|}{\textbf{Rad.}} & \textbf{Hau.} & \textbf{DTW} & \textbf{EDR} & \textbf{$|\Delta$Div.$|$} \\
        \midrule
        \\[-1.3em]
        \rowcolor{ivory2}
        \multicolumn{19}{l}{\textit{Without Road Network Graph}}                                                                                                                                                                                                                                                                                                                                                                                                                                                                                                                                         \\
        Dijkstra & 0.0026 & \multicolumn{1}{c|}{0.0029} & 0.5970 & 12.3210 & 0.3451 & 25.0 & 0.0161 & \multicolumn{1}{c|}{0.0074} & 0.4682 & 12.1164 & 0.3870 & 25.0 & 0.0068 & \multicolumn{1}{c|}{0.0035} & 0.4962 & 12.1404 & 0.4788 & 31.7 \\
        Markov & {\ul 0.0005} & \multicolumn{1}{c|}{{\ul 0.0003}} & 0.4761 & 9.0997 & 0.2453 & 37.4 & {\ul 0.0010} & \multicolumn{1}{c|}{0.0017} & 0.4645 & 12.4260 & 0.3276 & 46.7 & {\ul 0.0014} & \multicolumn{1}{c|}{0.0018} & 0.4655 & 11.9659 & 0.3844 & 46.2 \\
        MoveSim & 0.4927 & \multicolumn{1}{c|}{0.2220} & 12.6022 & 254.8898 & 0.9474 & 50.8 & 0.4152 & \multicolumn{1}{c|}{0.1423} & 4.0793 & 113.0140 & 0.9532 & 60.6 & 0.3117 & \multicolumn{1}{c|}{0.1854} & 3.6813 & 81.1215 & 0.9504 & 84.2 \\
        TrajSynVAE & 0.6923 & \multicolumn{1}{c|}{0.6856} & 15.5196 & 261.2315 & 0.9969 & N/A$^\dagger$ & 0.6825 & \multicolumn{1}{c|}{0.6252} & 5.8926 & 387.4879 & 0.9836 & N/A$^\dagger$ & 0.6879 & \multicolumn{1}{c|}{0.6266} & 7.4020 & 284.7321 & 0.9838 & N/A$^\dagger$ \\
        DiffTraj & 0.1348 & \multicolumn{1}{c|}{0.0030} & 1.2756 & 62.8626 & 0.8535 & 15.4 & 0.0313 & \multicolumn{1}{c|}{{\ul 0.0009}} & 0.4877 & 17.4727 & 0.5240 & 39.8 & 0.0336 & \multicolumn{1}{c|}{{\ul 0.0012}} & 0.7511 & 23.6415 & 0.7533 & 73.7 \\
        TrajFlow & 0.0041 & \multicolumn{1}{c|}{0.0039} & 1.1126 & 77.6046 & 0.5792 & {\ul 7.8} & 0.0121 & \multicolumn{1}{c|}{0.0033} & 0.7418 & 62.8708 & 0.6490 & {\ul 13.7} & 0.0092 & \multicolumn{1}{c|}{0.0040} & 0.8808 & 42.6706 & 0.5957 & 38.9 \\
        \midrule
        \\[-1.3em]
        \rowcolor{ivory2}
        \multicolumn{19}{l}{\textit{With Road Network Graph}}                                                                                                                                                                                                                                                                                                                                                                                                                                                                                                                                            \\
        TS-TrajGen & 0.0189 & \multicolumn{1}{c|}{0.0031} & 1.5167 & 40.4374 & 0.6190 & 28.5 & 0.0044 & \multicolumn{1}{c|}{0.0026} & 0.8428 & 25.8629 & 0.5958 & 23.3 & 0.0145 & \multicolumn{1}{c|}{0.0032} & 0.8699 & 25.3935 & 0.6980 & {\ul 18.8} \\
        STEGA & 0.1802 & \multicolumn{1}{c|}{0.0304} & 4.3107 & 141.8059 & 0.8709 & 164.3 & 0.1471 & \multicolumn{1}{c|}{0.0577} & 2.2545 & 253.7064 & 0.9636 & 14.7 & 0.6243 & \multicolumn{1}{c|}{0.1773} & 3.8843 & 593.0151 & 0.9638 & 84.7 \\
        HOSER & \textbf{0.0002} & \multicolumn{1}{c|}{\textbf{0.0001}} & \textbf{0.3554} & \textbf{5.9817} & \textbf{0.2061} & 23.3 & \textbf{0.0009} & \multicolumn{1}{c|}{\textbf{0.0004}} & {\ul 0.3065} & {\ul 7.4527} & {\ul 0.2383} & 26.6 & 0.0021 & \multicolumn{1}{c|}{0.0014} & {\ul 0.3659} & {\ul 8.4955} & {\ul 0.3359} & 29.9 \\
        \midrule
        TrajDLM & \textbf{0.0002} & \textbf{0.0001} & {\ul 0.3746} & {\ul 6.8438} & {\ul 0.2130} & \textbf{5.4} & {\ul 0.0010} & \textbf{0.0004} & \textbf{0.2375} & \textbf{5.5171} & \textbf{0.1911} & \textbf{7.0} & \textbf{0.0010} & \textbf{0.0008} & \textbf{0.3372} & \textbf{7.6713} & \textbf{0.3214} & \textbf{1.8} \\
        \bottomrule
    \end{tabular}}
\end{table*}

\subsection{Trajectory Rejection-Sampling Fine-Tuning}
\label{sec:rft}

Supervised training with TCS produces valid trajectories but tends to favor frequent, near-shortest routes, even when alternative samples for the same trip better match the ground truth (Appendix~\ref{app:rft-pseudocode}). TrajRFT addresses this limitation by fine-tuning the model on higher-quality sampled routes.

\paragraph{Sampling and Selection}

At each TrajRFT step, we sample a batch $\mathcal{B}$ of training trips, with each trip $\tau$ selected with probability proportional to its length $|\tau|$. Given a trip's prompt $x$ and ground-truth route $\tau$, the current model $G_\theta$ generates one greedy route $\hat{\tau}^{(0)}$ and $K$ sampled routes $\hat{\tau}^{(1)}, \dots, \hat{\tau}^{(K)}$ at temperature $\lambda = 1$, all under TCS. Among the sampled routes, we select the one closest to the ground truth in DTW distance, $k^\star = \arg\min_{k \geq 1} \mathrm{DTW}(\hat{\tau}^{(k)}, \tau)$, and retain it only if its DTW distance improves upon that of the greedy route by more than a relative margin $\epsilon$:

\begin{equation}
\frac{\mathrm{DTW}(\hat{\tau}^{(0)}, \tau) - \mathrm{DTW}(\hat{\tau}^{(k^\star)}, \tau)}{\mathrm{DTW}(\hat{\tau}^{(0)}, \tau)} > \epsilon.
\end{equation}

Because all candidate routes are generated under TCS, the selected training samples are topologically valid. The DTW-based selection further favors routes that better match the ground truth, rather than simply minimizing route length.

\paragraph{On-Policy Objective}
Let $\mathcal{S}$ denote the set of retained routes. For each retained route, we score its sampled segments $\hat{r}_i$ in the exact state from which they were generated, with the block containing position $i$ masked from $i$ onwards. We use the same constrained distribution as during sampling,
$\pi_\theta(\cdot \mid x, \hat{r}_{<i}) = \mathrm{Softmax}(\boldsymbol{f}_i + \boldsymbol{P}_{\hat{r}_{i-1},:} - \boldsymbol{D}_i)$,
so that training stays on-policy. Let $\pi_{\theta_0}$ denote the frozen supervised model. TrajRFT minimizes:
\begin{equation}
    \label{eq:rft}
    \begin{split}
        \mathcal{L}_{\mathrm{RFT}}(\theta) = {} & \frac{|\mathcal{S}|}{|\mathcal{B}|} \cdot \frac{1}{\sum_{\hat{\tau} \in \mathcal{S}} |J(\hat{\tau})|} \sum_{\hat{\tau} \in \mathcal{S}} \sum_{i \in J(\hat{\tau})} \Big[ -\log \pi_\theta(\hat{r}_i \mid x, \hat{r}_{<i}) \\
        & + \frac{\gamma}{2} \big( \log \pi_\theta(\hat{r}_i \mid x, \hat{r}_{<i}) - \log \pi_{\theta_0}(\hat{r}_i \mid x, \hat{r}_{<i}) \big)^2 \Big],
    \end{split}
\end{equation}
where the first term maximizes the likelihood of the selected routes and the second penalizes deviations from the supervised model using a $k_2$ estimate of the KL divergence. Here, $J(\hat{\tau})$ denotes the set of scored segments after the origin, comprising a random subset of $M$ segments for routes longer than $M$. Each scored segment receives equal weight, so longer retained routes contribute proportionally more to the objective. The pseudocode and the motivation for the design of  TrajRFT are provided in Appendix~\ref{app:rft-pseudocode}.

\subsection{TrajDLM: Overall Generation Framework}
\label{sec:generation}

Figure~\ref{fig:trajdlm-overview} illustrates the overall framework. Road segments are first encoded by the RNE to produce topology-aware embeddings, which replace the default token embeddings in the BDLM. The model is trained by minimizing the block-wise NELBO objective (Eq.~\ref{eq:bdnelbo}), where trajectories are decomposed into blocks and denoised via discrete diffusion conditioned on preceding blocks, and then fine-tuned with TrajRFT (Eq.~\ref{eq:rft}). At inference time, we generate a trajectory $\hat{\tau}$ given the conditioning tuple $(r_{\text{org}}, t_{\text{org}}, r_{\text{dest}})$ and trip attributes $(d_{\text{trip}}, \bar{d}_{\text{seg}}, t_{\text{trip}}, v_{\text{avg}})$. These conditions are converted into a textual prompt $x$, which is tokenized and provided as a prefix to $G_\theta$; the target is the trajectory's sequence of road segment tokens (Fig.~\ref{fig:trajdlm-prompt}). The trajectory sequence is initialized with mask tokens $\texttt{[M]}$. We then perform block-wise reverse diffusion, sampling each block as $\hat{\tau}^b \sim p_\theta(\tau^b \mid x, \hat{\tau}^{<b})$, applying TCS.

\begin{figure}[b]
    \centering
    \begin{tcolorbox}[promptstyle, title={Sample Input Prompt and Target Trajectory}]
        \footnotesize
        \textbf{Prompt:}

        Given the starting road segment \texttt{[RID\_24476]}, departure time 01:59, and ending road segment \texttt{[RID\_12280]}, along with the following attributes: trip distance 2491.76 meters, average distance between road segments 113.26 meters, trip time 5.83 minutes, and average speed 7.12 m/s, generate a plausible trajectory of road segments.

        \vspace{0.5em}

        \textbf{Target:}

        \texttt{[RID\_24476] [RID\_16147] [RID\_21210] [RID\_5626] [RID\_15719] [RID\_15662] [RID\_9652] [RID\_18361] [RID\_8939] [RID\_19809] [RID\_19434] [RID\_13676] [RID\_13683] [RID\_7621] [RID\_13685] [RID\_13687] [RID\_14596] [RID\_17255] [RID\_4814] [RID\_15575] [RID\_14967] [RID\_15579] [RID\_12280]}

    \end{tcolorbox}
    \caption{\textbf{Example of the input prompt and target trajectory sequence}. The prompt includes the origin and destination road segments, departure time, and trip-level attributes, while the target is a sequence of road segment IDs representing the trajectory to be generated.}
    \Description{A text box with two parts. The prompt is a sentence giving the starting road segment token, the departure time, the ending road segment token, and four trip attributes: trip distance, average distance between road segments, trip time, and average speed. The target is a sequence of 23 road segment ID tokens that begins with the starting segment and ends with the ending segment.}
    \label{fig:trajdlm-prompt}
\end{figure}

\section{Experiments}

\subsection{Experimental Setup}
\label{sec:experimental-setup}

\paragraph{Datasets}
\label{sec:datasets}

We train our model on three city-scale GPS trajectory datasets from Beijing, Porto\footnote{\url{https://www.kaggle.com/competitions/pkdd-15-taxi-trip-time-prediction-ii/data}}, and San Francisco\footnote{\url{https://ieee-dataport.org/open-access/crawdad-epflmobility}}, representing real-world mobility telemetry relevant to location-aware Web services and Web of Things applications. These datasets are adopted from HOSER, where raw GPS traces are map-matched~\cite{Yang2018FastMM} to road networks extracted from OpenStreetMap, converting continuous GPS coordinates into sequences of road segments. For fair comparison, we follow the same train/validation/test splits and use the publicly released preprocessed datasets from HOSER\footnote{\url{https://huggingface.co/datasets/caoji2001/HOSER-dataset}}. Additionally, we evaluate cross-domain generalization using GeoLife~\cite{zheng2010geolife}, which contains GPS trajectories recorded in Beijing from 2007--2012 using various GPS loggers and mobile phones across multiple transportation modes. Dataset statistics are provided in Appendix~\ref{app:datasets}.

\paragraph{Model}

We initialize our BDLM backbone from \nolinkurl{Qwen3-0.6B-diffusion-bd3lm-v0.1}\footnote{\url{https://huggingface.co/dllm-hub/Qwen3-0.6B-diffusion-bd3lm-v0.1}}, a variant of \texttt{Qwen3-0.6B}~\cite{qwen3technicalreport} adapted as a block diffusion language model via BD3-LM~\cite{arriola2025block,zhou2026dllm}. We set block length to $L' = 64$ for all cities. The Road Network Encoder uses a two-layer GATv2 with hidden dimension $d = 128$, followed by a two-layer MLP projection of hidden size $d_p = 512$ into the BDLM embedding space. At inference, we combine temperature-zero Gumbel-max sampling with TCS (Section~\ref{sec:tcd}). Additional implementation details, such as SFT, inference, and TrajRFT parameters, are provided in Appendix~\ref{app:implementation-details}.

\paragraph{Evaluation Metrics}

Adopting the evaluation framework used by HOSER and prior studies~\cite{cao2025hoser,jiang2023continuous,wang2024spatiotemporal}, we evaluate generated trajectories using both global and local metrics. For global metrics, we measure the Jensen-Shannon divergence (JSD) between the distributions of ground-truth and generated trajectories for two indicators: \ding{182} \emph{Distance} (total trip distance) and \ding{183} \emph{Radius} (radius of gyration~\cite{gonzalez2008understanding}). For local metrics, each generated trajectory is compared with the real trajectory it was generated for (same origin, destination, and trip attributes). Similarity is quantified with three standard trajectory similarity metrics: \ding{182} \emph{Hausdorff distance}~\cite{xie2017distributed}, \ding{183} \emph{DTW} (Dynamic Time Warping)~\cite{keogh2005exact}, and \ding{184} \emph{EDR} (Edit Distance on Real sequence)~\cite{chen2005robust}. We also compute \ding{185} \emph{Diversity}, which measures trajectory variety as the proportion of distinct road-segment $n$-grams among all $n$-grams across generated trajectories, using $n = 4$. Local similarity metrics alone may favor models that reproduce a limited set of routes without capturing the variability of real-world mobility. Conversely, high diversity does not necessarily imply realistic mobility, as random route generation can also produce diverse trajectories. We therefore additionally report $|\Delta$\textit{Div.}$|$ (\%): the absolute relative difference between the diversity of generated and ground-truth test trajectories, with lower values indicating closer agreement with real-world mobility patterns. All metrics are computed across all test trajectories.

\paragraph{Baselines}

We compare our model against a comprehensive suite of baselines, ranging from classical algorithms to state-of-the-art deep learning models. This includes \textbf{Classical Methods}: Dijkstra~\cite{dijkstra1959note}, Markov~\cite{gambs2012next}; \textbf{Generative Adversarial Networks (GANs)}: MoveSim~\cite{feng2020learning}, TS-TrajGen~\cite{jiang2023continuous}; \textbf{Variational Autoencoders (VAEs)}: TrajSynVAE~\cite{10.1109/TKDE.2023.3312209}; \textbf{Diffusion Models}: DiffTraj~\cite{zhu2023difftraj}; \textbf{Transformers}: STEGA~\cite{wang2024spatiotemporal}, HOSER~\cite{cao2025hoser}; and \textbf{Flow-Matching}: TrajFlow~\cite{li2026trajflow}. Following HOSER's protocol, we apply map-matching~\cite{Yang2018FastMM} to models that output continuous GPS coordinates prior to evaluation.

\subsection{Results}

\begin{table}[t]
    \centering
    \caption{\textbf{Trajectory generation performance on GeoLife}~\cite{zheng2010geolife}, overall and by transport mode. All models are trained on HOSER's Beijing dataset and evaluated in a zero-shot transfer setting on GeoLife. \textbf{Bold} marks the best score per mode; {\ul underline} marks the second-best.}
    \label{tab:geolife}
    \setlength{\tabcolsep}{2.5pt}
    \begin{tabular}{ll|ccc}
        \toprule
        \textbf{Mode} & \textbf{Method} & \textbf{Hau.} ($\downarrow$) & \textbf{DTW} ($\downarrow$) & \textbf{EDR} ($\downarrow$) \\
        \midrule
        \multirow{3}{*}{Unclassified} & Dijkstra & 0.5457 & 10.6348 & 0.4037 \\
         & HOSER & {\ul 0.4945} & \textbf{7.8716} & {\ul 0.3495} \\
         & TrajDLM & \textbf{0.4618} & {\ul 7.9839} & \textbf{0.3426} \\
        \midrule
        \multirow{3}{*}{Bike} & Dijkstra & \textbf{0.2484} & \textbf{3.6730} & \textbf{0.2668} \\
         & HOSER & {\ul 0.3570} & {\ul 5.0657} & {\ul 0.3786} \\
         & TrajDLM & 0.4158 & 6.5444 & 0.3863 \\
        \midrule
        \multirow{3}{*}{Walk} & Dijkstra & \textbf{0.1860} & \textbf{1.4825} & \textbf{0.3324} \\
         & HOSER & 0.3572 & {\ul 3.7036} & 0.3883 \\
         & TrajDLM & {\ul 0.2351} & 3.7297 & {\ul 0.3491} \\
        \midrule
        \multirow{3}{*}{Motorized} & Dijkstra & 0.9825 & 24.4530 & 0.5540 \\
         & HOSER & \textbf{0.7374} & {\ul 13.7345} & {\ul 0.4016} \\
         & TrajDLM & {\ul 0.7411} & \textbf{13.4291} & \textbf{0.3786} \\
        \midrule
        \multirow{3}{*}{All} & Dijkstra & 0.5295 & 10.3561 & 0.3977 \\
         & HOSER & {\ul 0.4894} & \textbf{7.7533} & {\ul 0.3545} \\
         & TrajDLM & \textbf{0.4609} & {\ul 7.9362} & \textbf{0.3469} \\
        \bottomrule
    \end{tabular}
\end{table}

\paragraph{Main Results}
\label{sec:main-results}

\begin{figure}[b]
    \centering
    \begin{subfigure}[t]{0.24\linewidth}
        \centering
        \includegraphics[width=\linewidth]{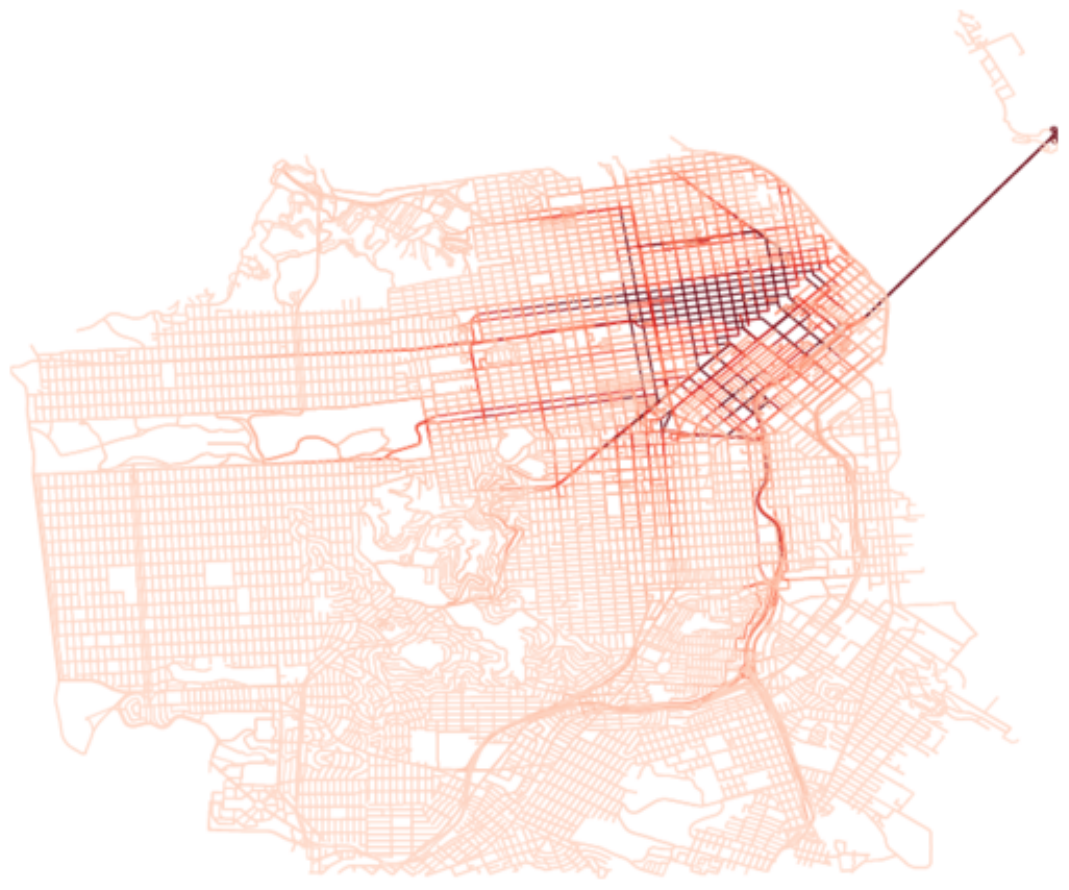}
        \caption{DiffTraj}
    \end{subfigure}
    \hfill
    \begin{subfigure}[t]{0.24\linewidth}
        \centering
        \includegraphics[width=\linewidth]{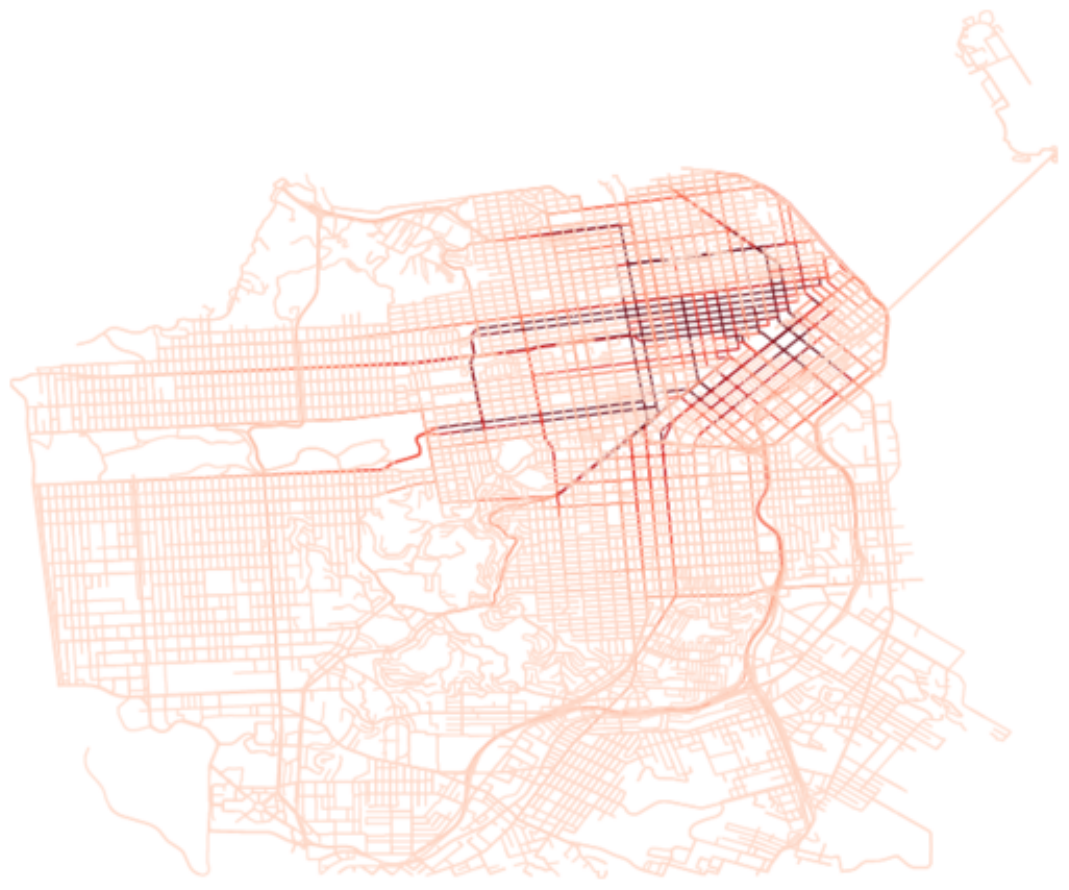}
        \caption{HOSER}
    \end{subfigure}
    \hfill
    \begin{subfigure}[t]{0.24\linewidth}
        \centering
        \includegraphics[width=\linewidth]{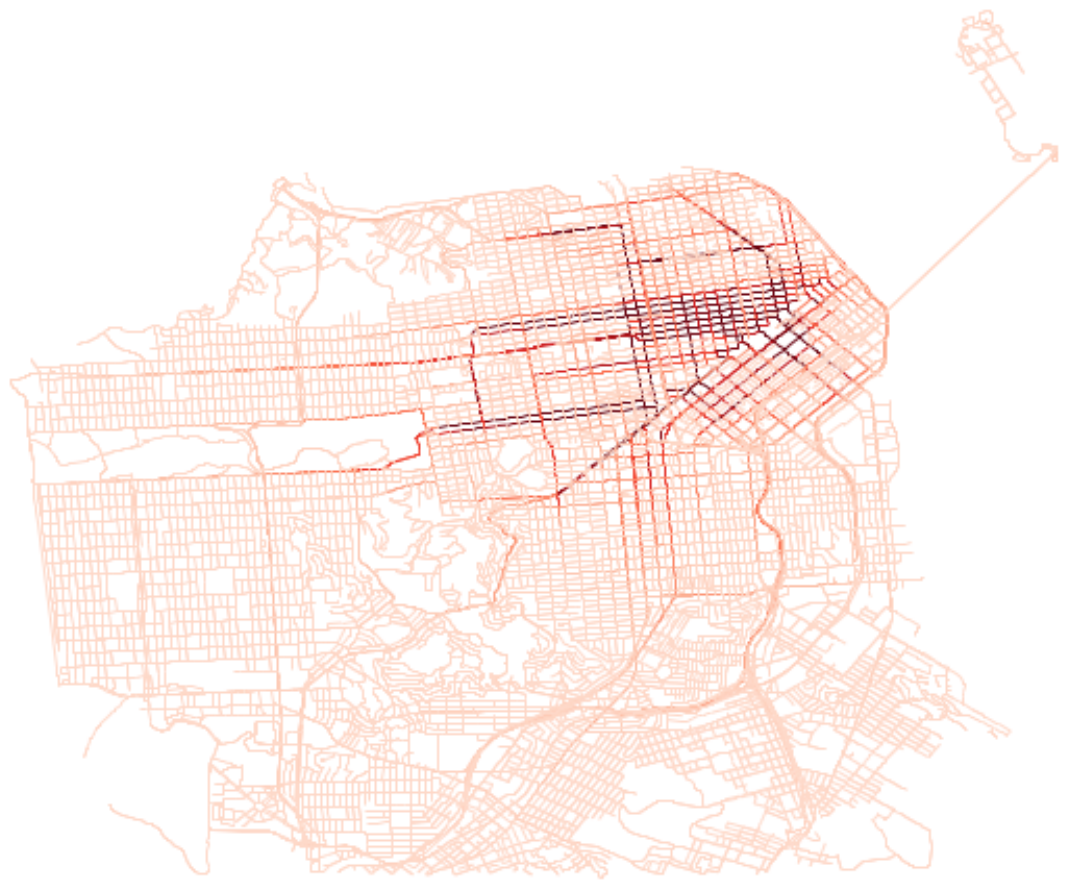}
        \caption{TrajDLM}
    \end{subfigure}
    \hfill
    \begin{subfigure}[t]{0.24\linewidth}
        \centering
        \includegraphics[width=\linewidth]{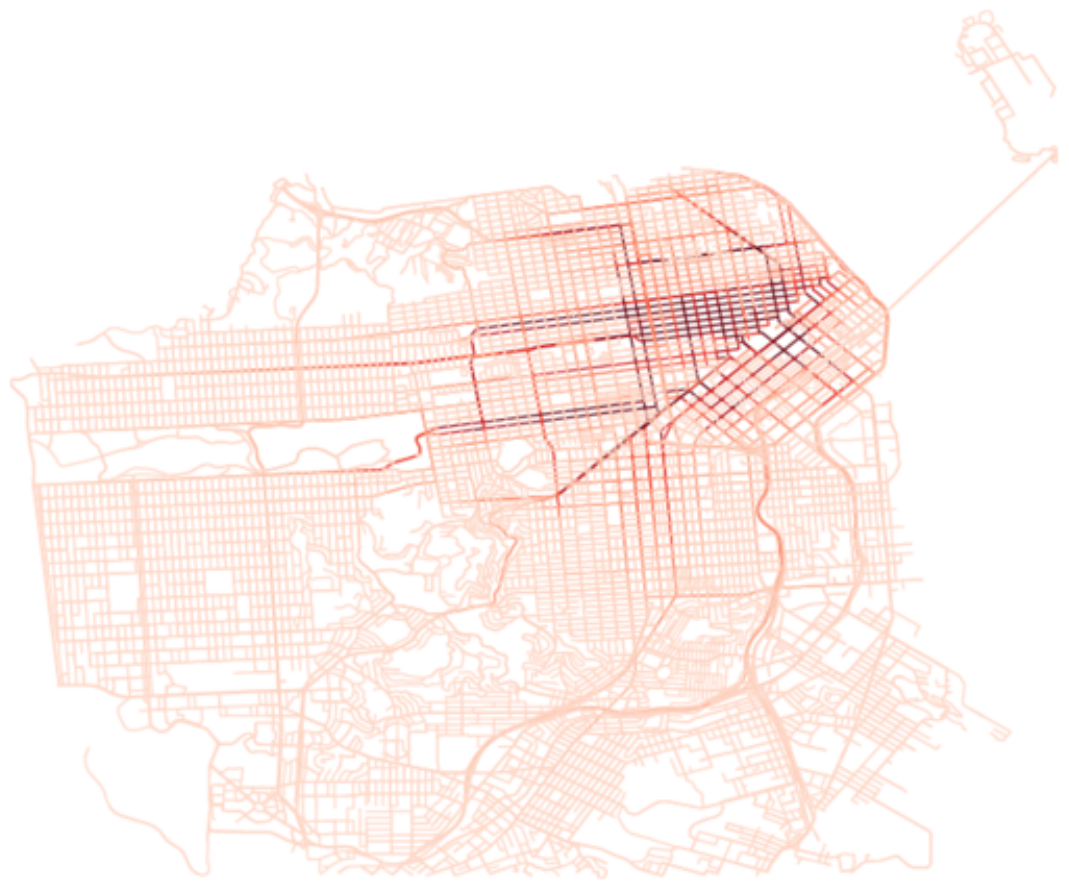}
        \caption{Real}
    \end{subfigure}
    \caption{\textbf{Trajectory density heatmaps on the San Francisco road network.} We compare DiffTraj, HOSER, and TrajDLM against real trajectories. Heatmap color indicates trajectory density. See Appendix~\ref{app:visualizations} for other visualizations.}
    \Description{Four side-by-side heatmaps of the San Francisco road network for DiffTraj, HOSER, TrajDLM, and real trajectories. Each road segment is colored by trajectory density, with darker red indicating more trajectories; the densest roads lie in the northeastern downtown area.}
    \label{fig:heatmap_comparison}
\end{figure}

Table~\ref{tab:main-results} compares TrajDLM against all baselines across the three datasets. Overall, TrajDLM achieves the best score on 14 of the 18 metrics and the second-best on the remaining 4. On Porto and San Francisco, it outperforms HOSER on every local metric, lowering \textit{DTW} by $26\%$ on Porto ($5.52$ vs.\ $7.45$) and $10\%$ on San Francisco ($7.67$ vs.\ $8.50$), and it is also best on both global metrics on San Francisco. The gains of TrajRFT concentrate on long trips, where the SFT model commits early to a wrong corridor and which account for most of its error (see Appendix~\ref{app:rft-pseudocode}). On Beijing, TrajDLM ties HOSER on both global metrics and is second to it on local metrics (\textit{DTW} $6.84$ vs.\ $5.98$). TrajDLM's routes are also the closest to real data in diversity on every city, within $5.4\%$, $7.0\%$, and $1.8\%$ of the real trajectories' diversity on Beijing, Porto, and San Francisco, compared with $23$--$30\%$ for HOSER, whose routes are less diverse than real routes. Compared to other baselines, continuous-space trajectory generation models like DiffTraj and TrajFlow generally perform worse as they do not explicitly enforce road network topology. Furthermore, consistent with~\cite{cao2025hoser}, classical approaches such as Markov and Dijkstra remain competitive on the global metrics, reflecting the tendency of real-world navigation to follow near-shortest paths~\cite{yuan2010t}, but fall well behind on the local metrics, as many real routes deviate from such paths. Qualitatively, Fig.~\ref{fig:heatmap_comparison} shows that TrajDLM produces trajectory heatmaps closely aligned with the real data distribution, whereas DiffTraj generates more spatially dispersed trajectories.

\begin{table*}[tbp]
    \centering
    \caption{\textbf{Downstream utility and generation efficiency.} \emph{Next-location prediction}: \textit{Acc@5} and \textit{MRR} of SASRec trained on synthetic trajectories and evaluated on real trajectories. \emph{Detour ratio}: median ratio of generated trajectory length to the shortest-path (Dijkstra) length between the same origin and destination. $\mu_{\text{s/traj}}$: average generation time per trajectory in seconds. \emph{Uniqueness}: median Hausdorff distance (km) to the nearest training trajectory~\cite{guo2026leveraging}; for Real, held-out test trajectories are used as the reference. \textbf{Bold} indicates the best result among synthetic data generators for each metric.}
    \label{tab:downstream}
    \setlength{\tabcolsep}{3pt}
    \resizebox{\textwidth}{!}{%
    \begin{tabular}{l|ccccc|ccccc|ccccc}
        \toprule
        & \multicolumn{5}{c|}{\textbf{Beijing}} & \multicolumn{5}{c|}{\textbf{Porto}} & \multicolumn{5}{c}{\textbf{San Francisco}} \\ \cmidrule{2-16}
        & \multicolumn{2}{c}{\textbf{Next-location} ($\uparrow$)} & \textbf{Route} ($\downarrow$) & \textbf{Eff.} ($\downarrow$) & \textbf{Privacy} & \multicolumn{2}{c}{\textbf{Next-location} ($\uparrow$)} & \textbf{Route} ($\downarrow$) & \textbf{Eff.} ($\downarrow$) & \textbf{Privacy} & \multicolumn{2}{c}{\textbf{Next-location} ($\uparrow$)} & \textbf{Route} ($\downarrow$) & \textbf{Eff.} ($\downarrow$) & \textbf{Privacy} \\
        \cmidrule(lr){2-3} \cmidrule(lr){4-4} \cmidrule(lr){5-5} \cmidrule(lr){6-6} \cmidrule(lr){7-8} \cmidrule(lr){9-9} \cmidrule(lr){10-10} \cmidrule(lr){11-11} \cmidrule(lr){12-13} \cmidrule(lr){14-14} \cmidrule(lr){15-15} \cmidrule(lr){16-16} 
        \textbf{Method} & \textbf{Acc@5} & \textbf{MRR} & \textbf{Detour} & $\mu_\textbf{s/traj}$ & \textbf{Uniq.} & \textbf{Acc@5} & \textbf{MRR} & \textbf{Detour} & $\mu_\textbf{s/traj}$ & \textbf{Uniq.} & \textbf{Acc@5} & \textbf{MRR} & \textbf{Detour} & $\mu_\textbf{s/traj}$ & \textbf{Uniq.} \\
        \midrule
        Dijkstra & 0.9741 & 0.9031 & -- & 0.004 & 0.268 & 0.9863 & 0.9220 & -- & 0.002 & 0.114 & 0.9688 & 0.8937 & -- & 0.005 & 0.183 \\
        Markov & 0.9723 & 0.9134 & 1.032 & \textbf{0.001} & 0.227 & 0.9851 & 0.9324 & 1.088 & \textbf{0.001} & 0.072 & 0.9680 & 0.9103 & 1.073 & \textbf{0.002} & 0.164 \\
        HOSER & 0.9805 & 0.9177 & 1.025 & 1.006 & 0.210 & 0.9920 & 0.9412 & 1.061 & 1.307 & 0.043 & 0.9755 & 0.9121 & \textbf{1.045} & 1.786 & 0.147 \\
        TrajDLM & \textbf{0.9844} & \textbf{0.9195} & \textbf{1.023} & 0.159 & 0.222 & \textbf{0.9950} & \textbf{0.9434} & \textbf{1.058} & 0.214 & 0.052 & \textbf{0.9815} & \textbf{0.9155} & \textbf{1.045} & 0.239 & 0.150 \\
        \midrule
        Real & 0.9901 & 0.9252 & -- & -- & 0.224 & 0.9973 & 0.9460 & -- & -- & 0.062 & 0.9882 & 0.9226 & -- & -- & 0.166 \\
        \bottomrule
    \end{tabular}}
\end{table*}

\begin{table}[bp]
    \centering
    \footnotesize
    \caption{\textbf{Robustness to imperfect preprocessing.} Local metrics on the first 5,000 test trajectories per city with perturbed inference-time inputs, compared with the clean ground truth.}
    \label{tab:robustness}
    \setlength{\tabcolsep}{3pt}
    \begin{tabular}{l|ccc}
        \toprule
        \textbf{Perturbation} & \textbf{Hau.} & \textbf{DTW} & \textbf{EDR} \\
        \midrule
        \rowcolor{ivory2} \multicolumn{4}{l}{\textit{Beijing}} \\
        None & 0.379 & 6.92 & 0.215 \\
        Attributes $\pm 20\%$ & 0.391 {\scriptsize(+3.1\%)} & 7.23 {\scriptsize(+4.4\%)} & 0.226 {\scriptsize(+5.0\%)} \\
        Destination snapped & 0.547 {\scriptsize(+44.3\%)} & 10.03 {\scriptsize(+44.8\%)} & 0.341 {\scriptsize(+58.9\%)} \\
        Both endpoints snapped & 0.752 {\scriptsize(+98.2\%)} & 14.98 {\scriptsize(+116.3\%)} & 0.489 {\scriptsize(+127.6\%)} \\
        1\% edges removed & 0.500 {\scriptsize(+31.9\%)} & 9.42 {\scriptsize(+36.1\%)} & 0.260 {\scriptsize(+21.1\%)} \\
        \midrule
        \rowcolor{ivory2} \multicolumn{4}{l}{\textit{Porto}} \\
        None & 0.241 & 5.73 & 0.193 \\
        Attributes $\pm 20\%$ & 0.286 {\scriptsize(+18.7\%)} & 6.82 {\scriptsize(+19.1\%)} & 0.235 {\scriptsize(+22.0\%)} \\
        Destination snapped & 0.292 {\scriptsize(+21.4\%)} & 6.46 {\scriptsize(+12.9\%)} & 0.260 {\scriptsize(+34.7\%)} \\
        Both endpoints snapped & 0.356 {\scriptsize(+47.9\%)} & 7.85 {\scriptsize(+37.0\%)} & 0.344 {\scriptsize(+78.6\%)} \\
        1\% edges removed & 0.534 {\scriptsize(+121.9\%)} & 14.27 {\scriptsize(+149.3\%)} & 0.348 {\scriptsize(+80.4\%)} \\
        \midrule
        \rowcolor{ivory2} \multicolumn{4}{l}{\textit{San Francisco}} \\
        None & 0.335 & 7.71 & 0.325 \\
        Attributes $\pm 20\%$ & 0.374 {\scriptsize(+11.5\%)} & 8.83 {\scriptsize(+14.5\%)} & 0.362 {\scriptsize(+11.5\%)} \\
        Destination snapped & 0.360 {\scriptsize(+7.5\%)} & 8.12 {\scriptsize(+5.4\%)} & 0.368 {\scriptsize(+13.3\%)} \\
        Both endpoints snapped & 0.431 {\scriptsize(+28.6\%)} & 10.04 {\scriptsize(+30.3\%)} & 0.471 {\scriptsize(+45.0\%)} \\
        1\% edges removed & 0.351 {\scriptsize(+4.8\%)} & 8.06 {\scriptsize(+4.6\%)} & 0.347 {\scriptsize(+6.9\%)} \\
        \bottomrule
    \end{tabular}
\end{table}

\begin{table*}[tbp]
    \centering
    \scriptsize
    \caption{\textbf{Ablation study} on the test split of each city. Each row removes or replaces one TrajDLM component. \emph{Valid}: fraction of generated trajectories that follow the road network. \textbf{Bold}: best value per column.}
    \label{tab:ablation}
    \setlength{\tabcolsep}{1.45pt}
    \resizebox{\textwidth}{!}{%
    \begin{tabular}{l|cc|cccc|c|cc|cccc|c|cc|cccc|c}
        \toprule
        & \multicolumn{7}{c|}{\textbf{Beijing}} & \multicolumn{7}{c|}{\textbf{Porto}} & \multicolumn{7}{c}{\textbf{San Francisco}} \\ \cmidrule{2-22}
        & \multicolumn{2}{c|}{\textbf{Global} ($\downarrow$)} & \multicolumn{4}{c|}{\textbf{Local} ($\downarrow$)} & \multicolumn{1}{c|}{\textbf{Topo.} ($\uparrow$)} & \multicolumn{2}{c|}{\textbf{Global} ($\downarrow$)} & \multicolumn{4}{c|}{\textbf{Local} ($\downarrow$)} & \multicolumn{1}{c|}{\textbf{Topo.} ($\uparrow$)} & \multicolumn{2}{c|}{\textbf{Global} ($\downarrow$)} & \multicolumn{4}{c|}{\textbf{Local} ($\downarrow$)} & \multicolumn{1}{c}{\textbf{Topo.} ($\uparrow$)} \\ \cmidrule{2-22}
        \textbf{Model} & \textbf{Dis.} & \textbf{Rad.} & \textbf{Hau.} & \textbf{DTW} & \textbf{EDR} & \textbf{$|\Delta$Div.$|$} & \textbf{Valid} & \textbf{Dis.} & \textbf{Rad.} & \textbf{Hau.} & \textbf{DTW} & \textbf{EDR} & \textbf{$|\Delta$Div.$|$} & \textbf{Valid} & \textbf{Dis.} & \textbf{Rad.} & \textbf{Hau.} & \textbf{DTW} & \textbf{EDR} & \textbf{$|\Delta$Div.$|$} & \textbf{Valid} \\
        \midrule
        w/o TrajRFT & 0.0013 & 0.0006 & 0.581 & 14.12 & 0.322 & 15.4 & 1.00 & 0.0018 & 0.0028 & 0.378 & 10.97 & 0.295 & 19.8 & 1.00 & 0.0018 & 0.0011 & 0.449 & 12.62 & 0.418 & 7.1 & 1.00 \\
        w/o TCS & 0.0097 & 0.0005 & 0.771 & 12.64 & 0.384 & 338.2 & 0.19 & 0.0013 & 0.0005 & 0.299 & 7.02 & 0.230 & 266.4 & 0.53 & \textbf{0.0008} & \textbf{0.0007} & 0.430 & 9.60 & 0.383 & 85.0 & 0.33 \\
        w/o RNE & 0.0799 & 0.0778 & 1.098 & 21.09 & 0.364 & 21.7 & 1.00 & \textbf{0.0008} & \textbf{0.0004} & \textbf{0.234} & 5.60 & \textbf{0.186} & 12.8 & 1.00 & 0.0028 & 0.0008 & 0.384 & 11.17 & 0.334 & 10.6 & 1.00 \\
        MDLM backbone & 0.0053 & 0.0054 & 0.393 & \textbf{6.77} & 0.227 & 6.1 & 1.00 & 0.0013 & 0.0005 & 0.279 & 6.26 & 0.226 & \textbf{3.5} & 1.00 & 0.0058 & 0.0054 & 0.413 & 9.79 & 0.376 & 5.2 & 1.00 \\
        \midrule
        TrajDLM & \textbf{0.0002} & \textbf{0.0001} & \textbf{0.375} & 6.84 & \textbf{0.213} & \textbf{5.4} & 1.00 & 0.0010 & \textbf{0.0004} & 0.238 & \textbf{5.52} & 0.191 & 7.0 & 1.00 & 0.0010 & 0.0008 & \textbf{0.337} & \textbf{7.67} & \textbf{0.321} & \textbf{1.8} & 1.00 \\
        \bottomrule
    \end{tabular}}
\end{table*}

\paragraph{Zero-Shot Cross-Domain Transfer}
\label{sec:geolife}

We evaluate the zero-shot cross-domain transferability of TrajDLM against HOSER and Dijkstra. Specifically, we transfer models trained on HOSER's taxi-based Beijing dataset (2015) to GeoLife trajectories collected in Beijing from 2007--2012, which cover diverse transportation modes~\cite{zheng2010geolife}. Table~\ref{tab:geolife} reports the local metrics both overall and by transportation mode, with \textit{Motorized} comprising car, bus, taxi, and subway trips. Overall, TrajDLM achieves the best \textit{Hausdorff} and \textit{EDR} scores and ranks a close second on \textit{DTW} ($7.94$ vs.\ $7.75$ for HOSER). TrajDLM ranks among the top two methods across all metrics for unclassified and motorized trips, which together account for $90\%$ of GeoLife trajectories. Dijkstra performs best only on bike and walk trips, which tend to follow near-shortest routes.

\subsection{Downstream Applications and Analysis}
\label{sec:downstream}

To evaluate TrajDLM as a synthetic trajectory generator, we examine four aspects: \ding{182} \emph{Next-location prediction}: downstream utility measured by prediction performance on real trajectories; \ding{183} \emph{Detour ratio}: route directness relative to shortest paths; \ding{184} \emph{Privacy}: closeness of generated trajectories to real trajectories; and \ding{185} \emph{Generation efficiency}: average generation time per trajectory. Table~\ref{tab:downstream} compares TrajDLM with Dijkstra, Markov, HOSER, and real data.

\paragraph{Next-Location Prediction}
We evaluate downstream utility using a train-on-synthetic, test-on-real protocol on each city's test split. We randomly divide the real test trajectories into an $80\%$ training subset and a $20\%$ evaluation subset, using the same split for all methods. For each method, we train a SASRec next-location prediction model~\cite{kang2018self} on synthetic trajectories generated for the training subset and evaluate it on the real trajectories in the evaluation subset, predicting the next road segment from the preceding segments. As a reference, we train the same model on the real training trajectories. Across all three cities, TrajDLM achieves the best next-location prediction performance among synthetic data generators on both \textit{Acc@5} and \textit{MRR}, closely approaching the real-data reference. HOSER performs second best, while classical baselines yield the lowest scores.

\paragraph{Detour Ratio}
To assess whether TrajDLM generates excessively detoured or unusual routes, we measure the detour ratio, defined as the generated trajectory length divided by the shortest-path length between the same origin and destination, computed using Dijkstra's algorithm. We report the median ratio across generated trajectories. TrajDLM achieves median detour ratios of $1.023$, $1.058$, and $1.045$ in Beijing, Porto, and San Francisco, respectively, indicating that generated routes are generally close to the shortest paths. These results suggest that TrajDLM captures route diversity without introducing excessive detours or unusual routes.

\paragraph{Throughput Efficiency}
We measure end-to-end batched throughput on the first 5,000 test trajectories per city, using a batch size of 16. TrajDLM generates each trajectory in $0.16$--$0.24$\,s, achieving up to $7.5\times$ higher throughput than HOSER despite its larger model size (0.6B vs.\ 2--5M parameters). This advantage stems from inference-time batching: TCS generates many trajectories in parallel, enabling efficient GPU utilization. In contrast, HOSER's A*-style search generates one road segment at a time for a single trajectory, making decoding inherently sequential and limiting effective GPU batching across trajectories. Shortest-path search is still significantly faster ($\leq 0.005$\,s), but at a large cost in fidelity (see Table~\ref{tab:main-results}). TrajDLM's higher batched throughput is valuable for mobility applications that require generating thousands of synthetic trajectories.

\paragraph{Privacy}
We assess trajectory memorization using the uniqueness test of~\cite{guo2026leveraging,de2013unique,yuan2024generating}, comparing each generated trajectory's nearest-training Hausdorff distance with that of 5,000 held-out real trips per city. TrajDLM's median distances ($0.222$, $0.052$, and $0.150$\,km for Beijing, Porto, and San Francisco) closely match the real-data references ($0.224$, $0.062$, and $0.166$\,km) and are closer to them than HOSER's ($0.210$, $0.043$, and $0.147$\,km), suggesting that its overlap with training routes reflects natural route recurrence rather than substantial additional memorization.

\paragraph{Robustness to Imperfect Preprocessing}
Map-matching GPS trajectories to a road network is standard practice in trajectory modeling, but can misassign endpoints or miss road connections, particularly with noisy traces or incomplete road networks. To assess robustness to such errors, we perturb TrajDLM's inference-time inputs on the first 5,000 test trajectories per city and compare the generated trajectories against clean ground truth (see Table~\ref{tab:robustness}). Perturbing all trip attributes by $\pm 20\%$ and snapping the destination to a nearby segment increase \textit{DTW} by $15$--$19\%$ and $5$--$13\%$, respectively, in Porto and San Francisco. In Porto, TrajDLM still outperforms HOSER with clean inputs (see Table~\ref{tab:main-results}). Perturbations to inputs enforced as hard constraints by TCS have a greater impact. Snapping both endpoints increases \textit{DTW} by $30$--$37\%$ in Porto and San Francisco. Removing $1\%$ of road-network edges, which disconnects $24$--$40\%$ of real routes, increases \textit{DTW} by $5\%$ in San Francisco, $36\%$ in Beijing, and $149\%$ in Porto, where detours around missing edges are longer. Beijing is also particularly sensitive to endpoint errors: a snapped endpoint often falls on the same road in the opposite direction, requiring an average road detour of $510$\,m, compared with $170$--$213$\,m in Porto and San Francisco. These findings highlight a limitation of TCS: although it guarantees valid routes between the specified endpoints on the supplied road graph, it cannot correct errors in either input. Accurate map-matching and road-network data remain important.

\subsection{Ablation Study}
\label{sec:ablation}

Table~\ref{tab:ablation} ablates each component of TrajDLM through four variants: removing TrajRFT (w/o TrajRFT), replacing topology-constrained sampling with the default BD3-LM sampler (w/o TCS), removing the road network encoder and using vanilla token embeddings (w/o RNE), and replacing the block diffusion backbone with a masked diffusion language model (MDLM)~\cite{sahoo2024simple}, which denoises the full sequence jointly rather than block by block. It is initialized from \texttt{Qwen3-0.6B-diffusion-mdlm-v0.1}~\cite{zhou2026dllm}, the MDLM counterpart of our backbone, and trained with the same SFT, TrajRFT, and TCS (MDLM backbone).

\textbf{TrajRFT.} The second training stage gives a large gain in local fidelity, lowering \textit{DTW} by $52\%$ on Beijing ($14.12 \rightarrow 6.84$), $50\%$ on Porto ($10.97 \rightarrow 5.52$), and $39\%$ on San Francisco ($12.62 \rightarrow 7.67$), with smaller but consistent gains in \textit{Hausdorff} and \textit{EDR}.

\textbf{TCS.} Removing TCS at inference time reduces trajectory validity to $19\%$, $53\%$, and $33\%$ on Beijing, Porto, and San Francisco, while \textit{DTW} rises to $12.64$, $7.02$, and $9.60$. Although TrajRFT keeps fine-tuning on-policy through TCS-based sampling, TCS remains necessary at inference time to guarantee topological validity.

\textbf{RNE.} Replacing the RNE with vanilla token embeddings increases \textit{DTW} from $6.84$ to $21.09$ on Beijing and from $7.67$ to $11.17$ on San Francisco. Porto shows little change ($5.52$ vs.\ $5.60$), showing that GNN-derived, topology-aware road representations improve route generation fidelity, particularly on Beijing and San Francisco.

\textbf{SFT.} Without SFT, the pretrained BDLM generates no trajectories under TCS, emitting only empty reasoning blocks and end-of-sequence tokens at both greedy and sampling temperatures. Across the 3 cities and 100 TrajRFT steps, no candidate is retained, leaving the model without a training signal. SFT is therefore necessary to initialize a policy that generates routes for TrajRFT to improve.

\textbf{MDLM Backbone.} Replacing block diffusion with MDLM generally reduces fidelity. On San Francisco and Porto, \textit{DTW} increases from $7.67$ to $9.79$ and from $5.52$ to $6.26$, respectively. On Beijing, MDLM achieves comparable \textit{DTW} ($6.77$ vs.\ $6.84$) but performs worse on \textit{Hausdorff}, \textit{EDR}, and both global metrics. MDLM better matches real-data diversity only on Porto ($3.5\%$ vs.\ $7.0\%$).

\section{Conclusion}

This work introduces TrajDLM, a topology-aware block diffusion language model for generating synthetic mobility telemetry. By combining graph-based road network representations with block-wise denoising, TrajDLM generates realistic trajectories over discrete road segments, while topology-constrained sampling guarantees valid road transitions and preserves specified origins and destinations. To improve route fidelity beyond supervised training, TrajRFT fine-tunes the model on self-generated routes selected for their similarity to real trajectories. Experiments across three city-scale datasets show that TrajDLM captures population-level mobility distributions, individual route patterns, and route diversity, while generating trajectories up to $7.5\times$ faster than the strongest baseline. It also transfers to unseen domains in a zero-shot setting, remaining competitive overall and on motorized trips. These results establish TrajDLM as an efficient approach to generating realistic, topologically valid mobility telemetry for location-aware Web services and Web of Things applications.

\begin{acks}
We would like to thank the support of the ARC Center of Excellence for Automated Decision Making and Society (CE200100005). We express our gratitude to Sharon AI for providing access to NVIDIA H100 GPUs.
\end{acks}

\bibliographystyle{ACM-Reference-Format}
\bibliography{References}

@inproceedings{cao2025hoser,
  title     = {Holistic Semantic Representation for Navigational Trajectory Generation},
  author    = {Cao, Ji and Zheng, Tongya and Guo, Qinghong and Wang, Yu and Dai, Junshu and Liu, Shunyu and Yang, Jie and Song, Jie and Song, Mingli},
  booktitle = {Proceedings of the AAAI Conference on Artificial Intelligence},
  volume    = {39},
  number    = {1},
  pages     = {40--48},
  year      = {2025}
}

@article{zhu2023difftraj,
  title   = {Difftraj: Generating gps trajectory with diffusion probabilistic model},
  author  = {Zhu, Yuanshao and Ye, Yongchao and Zhang, Shiyao and Zhao, Xiangyu and Yu, James},
  journal = {Advances in Neural Information Processing Systems},
  volume  = {36},
  pages   = {65168--65188},
  year    = {2023}
}

@inproceedings{arriola2025block,
  title     = {Block Diffusion: Interpolating Between Autoregressive and Diffusion Language Models},
  author    = {Marianne Arriola and Subham Sekhar Sahoo and Aaron Gokaslan and Zhihan Yang and Zhixuan Qi and Jiaqi Han and Justin T Chiu and Volodymyr Kuleshov},
  booktitle = {The Thirteenth International Conference on Learning Representations},
  year      = {2025},
  url       = {https://openreview.net/forum?id=tyEyYT267x}
}

@inproceedings{sohl2015deep,
  title        = {Deep unsupervised learning using nonequilibrium thermodynamics},
  author       = {Sohl-Dickstein, Jascha and Weiss, Eric and Maheswaranathan, Niru and Ganguli, Surya},
  booktitle    = {International conference on machine learning},
  pages        = {2256--2265},
  year         = {2015},
  organization = {pmlr}
}

@article{austin2021structured,
  title   = {Structured denoising diffusion models in discrete state-spaces},
  author  = {Austin, Jacob and Johnson, Daniel D and Ho, Jonathan and Tarlow, Daniel and Van Den Berg, Rianne},
  journal = {Advances in neural information processing systems},
  volume  = {34},
  pages   = {17981--17993},
  year    = {2021}
}

@article{Yang2018FastMM,
  title   = {Fast map matching, an algorithm integrating hidden Markov model with precomputation},
  author  = {Can Yang and Gyozo Gidofalvi},
  journal = {International Journal of Geographical Information Science},
  year    = {2018},
  volume  = {32},
  number  = {3},
  pages   = {547 - 570}
}

@inproceedings{jiang2023continuous,
  title     = {Continuous trajectory generation based on two-stage GAN},
  author    = {Jiang, Wenjun and Zhao, Wayne Xin and Wang, Jingyuan and Jiang, Jiawei},
  booktitle = {Proceedings of the AAAI conference on artificial intelligence},
  volume    = {37},
  number    = {4},
  pages     = {4374--4382},
  year      = {2023}
}

@article{wang2024spatiotemporal,
  title     = {Spatiotemporal gated traffic trajectory simulation with semantic-aware graph learning},
  author    = {Wang, Yu and Cao, Ji and Huang, Wenjie and Liu, Zhihua and Zheng, Tongya and Song, Mingli},
  journal   = {Information Fusion},
  volume    = {108},
  pages     = {102404},
  year      = {2024},
  publisher = {Elsevier}
}

@article{gonzalez2008understanding,
  title     = {Understanding individual human mobility patterns},
  author    = {Gonzalez, Marta C and Hidalgo, Cesar A and Barabasi, Albert-Laszlo},
  journal   = {nature},
  volume    = {453},
  number    = {7196},
  pages     = {779--782},
  year      = {2008},
  publisher = {Nature Publishing Group UK London}
}

@article{xie2017distributed,
  title     = {Distributed trajectory similarity search},
  author    = {Xie, Dong and Li, Feifei and Phillips, Jeff M},
  journal   = {Proceedings of the VLDB Endowment},
  volume    = {10},
  number    = {11},
  pages     = {1478--1489},
  year      = {2017},
  publisher = {VLDB Endowment}
}

@article{keogh2005exact,
  title     = {Exact indexing of dynamic time warping},
  author    = {Keogh, Eamonn and Ratanamahatana, Chotirat Ann},
  journal   = {Knowledge and information systems},
  volume    = {7},
  number    = {3},
  pages     = {358--386},
  year      = {2005},
  publisher = {Springer}
}

@inproceedings{chen2005robust,
  title     = {Robust and fast similarity search for moving object trajectories},
  author    = {Chen, Lei and {\"O}zsu, M Tamer and Oria, Vincent},
  booktitle = {Proceedings of the 2005 ACM SIGMOD international conference on Management of data},
  pages     = {491--502},
  year      = {2005}
}

@inproceedings{gambs2012next,
  title     = {Next place prediction using mobility markov chains},
  author    = {Gambs, S{\'e}bastien and Killijian, Marc-Olivier and del Prado Cortez, Miguel N{\'u}{\~n}ez},
  booktitle = {Proceedings of the first workshop on measurement, privacy, and mobility},
  pages     = {1--6},
  year      = {2012}
}

@article{dijkstra1959note,
  title   = {A note on two problems in connexion with graphs},
  author  = {Dijkstra, EW},
  journal = {Numerische Mathematik},
  volume  = {1},
  number  = {1},
  pages   = {269--271},
  year    = {1959}
}

@inproceedings{yu2017seqgan,
  title     = {Seqgan: Sequence generative adversarial nets with policy gradient},
  author    = {Yu, Lantao and Zhang, Weinan and Wang, Jun and Yu, Yong},
  booktitle = {Proceedings of the AAAI conference on artificial intelligence},
  volume    = {31},
  number    = {1},
  year      = {2017}
}

@inproceedings{huang2019variational,
  title        = {A variational autoencoder based generative model of urban human mobility},
  author       = {Huang, Dou and Song, Xuan and Fan, Zipei and Jiang, Renhe and Shibasaki, Ryosuke and Zhang, Yu and Wang, Haizhong and Kato, Yugo},
  booktitle    = {2019 IEEE conference on multimedia information processing and retrieval (MIPR)},
  pages        = {425--430},
  year         = {2019},
  organization = {IEEE}
}

@inproceedings{feng2020learning,
  title     = {Learning to simulate human mobility},
  author    = {Feng, Jie and Yang, Zeyu and Xu, Fengli and Yu, Haisu and Wang, Mudan and Li, Yong},
  booktitle = {Proceedings of the 26th ACM SIGKDD international conference on knowledge discovery \& data mining},
  pages     = {3426--3433},
  year      = {2020}
}

@inproceedings{cao2021generating,
  title     = {Generating mobility trajectories with retained data utility},
  author    = {Cao, Chu and Li, Mo},
  booktitle = {Proceedings of the 27th ACM SIGKDD conference on knowledge discovery \& data mining},
  pages     = {2610--2620},
  year      = {2021}
}

@inproceedings{brody2022how,
  title     = {How Attentive are Graph Attention Networks? },
  author    = {Shaked Brody and Uri Alon and Eran Yahav},
  booktitle = {International Conference on Learning Representations},
  year      = {2022},
  url       = {https://openreview.net/forum?id=F72ximsx7C1}
}

@article{zheng2010geolife,
  title   = {GeoLife: A collaborative social networking service among user, location and trajectory.},
  author  = {Zheng, Yu and Xie, Xing and Ma, Wei-Ying and others},
  journal = {IEEE Data Eng. Bull.},
  volume  = {33},
  number  = {2},
  pages   = {32--39},
  year    = {2010}
}

@article{10.1109/TKDE.2023.3312209,
  author     = {Wang, Huandong and Zhang, Qizhong and Wu, Yuchen and Jin, Depeng and Wang, Xing and Zhu, Lin and Yu, Li},
  title      = {Synthesizing Human Trajectories Based on Variational Point Processes},
  year       = {2024},
  issue_date = {April 2024},
  publisher  = {IEEE Educational Activities Department},
  address    = {USA},
  volume     = {36},
  number     = {4},
  issn       = {1041-4347},
  url        = {https://doi.org/10.1109/TKDE.2023.3312209},
  doi        = {10.1109/TKDE.2023.3312209},
  journal    = {IEEE Trans. on Knowl. and Data Eng.},
  month      = apr,
  pages      = {1785–1799},
  numpages   = {15}
}

@inproceedings{li2026trajflow,
  title     = {TrajFlow: Nation-wide Pseudo {GPS} Trajectory Generation with Flow Matching Models},
  author    = {Peiran Li and Jiawei Wang and Haoran Zhang and Xiaodan Shi and Noboru Koshizuka and Chihiro Shimizu and Renhe Jiang},
  booktitle = {The Fourteenth International Conference on Learning Representations},
  year      = {2026},
  url       = {https://openreview.net/forum?id=BDOldEjwCE}
}

@misc{qwen3technicalreport,
  title         = {Qwen3 Technical Report},
  author        = {Qwen Team},
  year          = {2025},
  eprint        = {2505.09388},
  archiveprefix = {arXiv},
  primaryclass  = {cs.CL},
  url           = {https://arxiv.org/abs/2505.09388}
}

@misc{zhou2026dllm,
  title         = {dLLM: Simple Diffusion Language Modeling},
  author        = {Zhanhui Zhou and Lingjie Chen and Hanghang Tong and Dawn Song},
  year          = {2026},
  eprint        = {2602.22661},
  archiveprefix = {arXiv},
  primaryclass  = {cs.CL},
  url           = {https://arxiv.org/abs/2602.22661}
}

@inproceedings{yuan2010t,
  title     = {T-drive: driving directions based on taxi trajectories},
  author    = {Yuan, Jing and Zheng, Yu and Zhang, Chengyang and Xie, Wenlei and Xie, Xing and Sun, Guangzhong and Huang, Yan},
  booktitle = {Proceedings of the 18th SIGSPATIAL International conference on advances in geographic information systems},
  pages     = {99--108},
  year      = {2010}
}

@article{zheng2023spatial,
  title     = {Spatial planning of urban communities via deep reinforcement learning},
  author    = {Zheng, Yu and Lin, Yuming and Zhao, Liang and Wu, Tinghai and Jin, Depeng and Li, Yong},
  journal   = {Nature Computational Science},
  volume    = {3},
  number    = {9},
  pages     = {748--762},
  year      = {2023},
  publisher = {Nature Publishing Group US New York}
}

@inproceedings{zheng2023road,
  title     = {Road planning for slums via deep reinforcement learning},
  author    = {Zheng, Yu and Su, Hongyuan and Ding, Jingtao and Jin, Depeng and Li, Yong},
  booktitle = {Proceedings of the 29th ACM SIGKDD Conference on Knowledge Discovery and Data Mining},
  pages     = {5695--5706},
  year      = {2023}
}

@inproceedings{jiang2016route,
  title     = {Route planning for locations based on trajectory segments},
  author    = {Jiang, Jinsheng and Xu, Chong and Xu, Jian and Xu, Ming and Zheng, Ning and Kong, Kaiwei},
  booktitle = {Proceedings of the 2nd ACM SIGSPATIAL Workshop on Smart Cities and Urban Analytics},
  pages     = {1--8},
  year      = {2016}
}

@article{lv2014traffic,
  title     = {Traffic flow prediction with big data: A deep learning approach},
  author    = {Lv, Yisheng and Duan, Yanjie and Kang, Wenwen and Li, Zhengxi and Wang, Fei-Yue},
  journal   = {Ieee transactions on intelligent transportation systems},
  volume    = {16},
  number    = {2},
  pages     = {865--873},
  year      = {2014},
  publisher = {IEEE}
}

@article{jin2023spatio,
  title     = {Spatio-temporal graph neural networks for predictive learning in urban computing: A survey},
  author    = {Jin, Guangyin and Liang, Yuxuan and Fang, Yuchen and Shao, Zezhi and Huang, Jincai and Zhang, Junbo and Zheng, Yu},
  journal   = {IEEE transactions on knowledge and data engineering},
  volume    = {36},
  number    = {10},
  pages     = {5388--5408},
  year      = {2023},
  publisher = {IEEE}
}

@article{de2013unique,
  title     = {Unique in the crowd: The privacy bounds of human mobility},
  author    = {De Montjoye, Yves-Alexandre and Hidalgo, C{\'e}sar A and Verleysen, Michel and Blondel, Vincent D},
  journal   = {Scientific reports},
  volume    = {3},
  number    = {1},
  pages     = {1376},
  year      = {2013},
  publisher = {Nature Publishing Group UK London}
}

@article{chen2024trajectory,
  title   = {Trajectory Data Management and Mining: A Survey from Deep Learning to the LLM Era},
  author  = {Chen, Wei and Zhu, Yuanshao and Chang, Yanchuan and Luo, Kang and Wen, Haomin and Li, Lei and Yu, Yanwei and Wen, Qingsong and Chen, Chao and Zheng, Kai and others},
  journal = {arXiv preprint arXiv:2403.14151},
  year    = {2024}
}

@article{yuan2025breaking,
  title   = {Breaking Data Silos: Towards Open and Scalable Mobility Foundation Models via Generative Continual Learning},
  author  = {Yuan, Yuan and Liu, Yukun and Han, Chonghua and Feng, Jie and Li, Yong},
  journal = {arXiv preprint arXiv:2506.06694},
  year    = {2025}
}

@article{yang2023diffusion,
  title     = {Diffusion models: A comprehensive survey of methods and applications},
  author    = {Yang, Ling and Zhang, Zhilong and Song, Yang and Hong, Shenda and Xu, Runsheng and Zhao, Yue and Zhang, Wentao and Cui, Bin and Yang, Ming-Hsuan},
  journal   = {ACM computing surveys},
  volume    = {56},
  number    = {4},
  pages     = {1--39},
  year      = {2023},
  publisher = {ACM New York, NY, USA}
}

@inproceedings{10.1145/3394486.3403043,
  author    = {Wu, Ning and Zhao, Xin Wayne and Wang, Jingyuan and Pan, Dayan},
  title     = {Learning Effective Road Network Representation with Hierarchical Graph Neural Networks},
  year      = {2020},
  isbn      = {9781450379984},
  publisher = {Association for Computing Machinery},
  address   = {New York, NY, USA},
  url       = {https://doi.org/10.1145/3394486.3403043},
  doi       = {10.1145/3394486.3403043},
  booktitle = {Proceedings of the 26th ACM SIGKDD International Conference on Knowledge Discovery \& Data Mining},
  pages     = {6–14},
  numpages  = {9},
  location  = {Virtual Event, CA, USA},
  series    = {KDD '20}
}

@inproceedings{10.1145/3347146.3359094,
  author    = {Jepsen, Tobias Skovgaard and Jensen, Christian S. and Nielsen, Thomas Dyhre},
  title     = {Graph Convolutional Networks for Road Networks},
  year      = {2019},
  isbn      = {9781450369091},
  publisher = {Association for Computing Machinery},
  address   = {New York, NY, USA},
  url       = {https://doi.org/10.1145/3347146.3359094},
  doi       = {10.1145/3347146.3359094},
  booktitle = {Proceedings of the 27th ACM SIGSPATIAL International Conference on Advances in Geographic Information Systems},
  pages     = {460–463},
  numpages  = {4},
  location  = {Chicago, IL, USA},
  series    = {SIGSPATIAL '19}
}

@article{sahoo2024simple,
  title   = {Simple and effective masked diffusion language models},
  author  = {Sahoo, Subham S and Arriola, Marianne and Schiff, Yair and Gokaslan, Aaron and Marroquin, Edgar and Chiu, Justin T and Rush, Alexander and Kuleshov, Volodymyr},
  journal = {Advances in Neural Information Processing Systems},
  volume  = {37},
  pages   = {130136--130184},
  year    = {2024}
}

@article{nie2025large,
  title   = {Large language diffusion models},
  author  = {Nie, Shen and Zhu, Fengqi and You, Zebin and Zhang, Xiaolu and Ou, Jingyang and Hu, Jun and Zhou, Jun and Lin, Yankai and Wen, Ji-Rong and Li, Chongxuan},
  journal = {arXiv preprint arXiv:2502.09992},
  year    = {2025}
}

@inproceedings{zhu2024controltraj,
  title     = {Controltraj: Controllable trajectory generation with topology-constrained diffusion model},
  author    = {Zhu, Yuanshao and Yu, James Jianqiao and Zhao, Xiangyu and Liu, Qidong and Ye, Yongchao and Chen, Wei and Zhang, Zijian and Wei, Xuetao and Liang, Yuxuan},
  booktitle = {Proceedings of the 30th ACM SIGKDD Conference on Knowledge Discovery and Data Mining},
  pages     = {4676--4687},
  year      = {2024}
}

@inproceedings{zhuunitraj,
  title     = {UniTraj: Learning a Universal Trajectory Foundation Model from Billion-Scale Worldwide Traces},
  author    = {Zhu, Yuanshao and Yu, James Jianqiao and Zhao, Xiangyu and Zhou, Xun and Han, Liang and Wei, Xuetao and Liang, Yuxuan},
  booktitle = {The Thirty-ninth Annual Conference on Neural Information Processing Systems},
  year      = {2025}
}

@inproceedings{liu2018trajgans,
  title     = {trajGANs: Using generative adversarial networks for geo-privacy protection of trajectory data (Vision paper)},
  author    = {Liu, Xia and Chen, Hanzhou and Andris, Clio},
  booktitle = {Location privacy and security workshop},
  pages     = {1--7},
  year      = {2018}
}

@article{zhang2023dp,
  title     = {DP-TrajGAN: A privacy-aware trajectory generation model with differential privacy},
  author    = {Zhang, Jing and Huang, Qihan and Huang, Yirui and Ding, Qian and Tsai, Pei-Wei},
  journal   = {Future Generation Computer Systems},
  volume    = {142},
  pages     = {25--40},
  year      = {2023},
  publisher = {Elsevier}
}

@inproceedings{li2024t,
  title     = {T-jepa: A joint-embedding predictive architecture for trajectory similarity computation},
  author    = {Li, Lihuan and Xue, Hao and Song, Yang and Salim, Flora},
  booktitle = {Proceedings of the 32nd ACM international conference on advances in geographic information systems},
  pages     = {569--572},
  year      = {2024}
}

@article{li2025hit,
  title   = {HiT-JEPA: A Hierarchical Self-supervised Trajectory Embedding Framework for Similarity Computation},
  author  = {Li, Lihuan and Xue, Hao and Ao, Shuang and Song, Yang and Salim, Flora},
  journal = {arXiv preprint arXiv:2507.00028},
  year    = {2025}
}

@inproceedings{deng2025marionette,
  title     = {Marionette: Fine-grained conditional generative modeling of spatiotemporal human trajectory data beyond imitation},
  author    = {Deng, Bangchao and Ding, Ling and Ji, Lianhua and Chen, Chunhua and Jing, Xin and Qu, Bingqing and Yang, Dingqi},
  booktitle = {Proceedings of the 31st ACM SIGKDD Conference on Knowledge Discovery and Data Mining V. 2},
  pages     = {463--473},
  year      = {2025}
}

@inproceedings{chang2023contrastive,
  title        = {Contrastive trajectory similarity learning with dual-feature attention},
  author       = {Chang, Yanchuan and Qi, Jianzhong and Liang, Yuxuan and Tanin, Egemen},
  booktitle    = {2023 IEEE 39th International conference on data engineering (ICDE)},
  pages        = {2933--2945},
  year         = {2023},
  organization = {IEEE}
}

@article{kipf2016semi,
  title   = {Semi-supervised classification with graph convolutional networks},
  author  = {Kipf, Thomas N and Welling, Max},
  journal = {arXiv preprint arXiv:1609.02907},
  year    = {2016}
}

@article{kong2020diffwave,
  title   = {Diffwave: A versatile diffusion model for audio synthesis},
  author  = {Kong, Zhifeng and Ping, Wei and Huang, Jiaji and Zhao, Kexin and Catanzaro, Bryan},
  journal = {arXiv preprint arXiv:2009.09761},
  year    = {2020}
}

@inproceedings{rombach2022high,
  title     = {High-resolution image synthesis with latent diffusion models},
  author    = {Rombach, Robin and Blattmann, Andreas and Lorenz, Dominik and Esser, Patrick and Ommer, Bj{\"o}rn},
  booktitle = {Proceedings of the IEEE/CVF conference on computer vision and pattern recognition},
  pages     = {10684--10695},
  year      = {2022}
}

@article{qin2023diffusion,
  title     = {A diffusion model for poi recommendation},
  author    = {Qin, Yifang and Wu, Hongjun and Ju, Wei and Luo, Xiao and Zhang, Ming},
  journal   = {ACM Transactions on Information Systems},
  volume    = {42},
  number    = {2},
  pages     = {1--27},
  year      = {2023},
  publisher = {ACM New York, NY}
}

@article{li2025survey,
  title   = {A survey on diffusion language models},
  author  = {Li, Tianyi and Chen, Mingda and Guo, Bowei and Shen, Zhiqiang},
  journal = {arXiv preprint arXiv:2508.10875},
  year    = {2025}
}

@article{ye2025dream,
  title   = {Dream 7b: Diffusion large language models},
  author  = {Ye, Jiacheng and Xie, Zhihui and Zheng, Lin and Gao, Jiahui and Wu, Zirui and Jiang, Xin and Li, Zhenguo and Kong, Lingpeng},
  journal = {arXiv preprint arXiv:2508.15487},
  year    = {2025}
}

@article{bie2025llada2,
  title   = {Llada2. 0: Scaling up diffusion language models to 100b},
  author  = {Bie, Tiwei and Cao, Maosong and Chen, Kun and Du, Lun and Gong, Mingliang and Gong, Zhuochen and Gu, Yanmei and Hu, Jiaqi and Huang, Zenan and Lan, Zhenzhong and others},
  journal = {arXiv preprint arXiv:2512.15745},
  year    = {2025}
}

@article{bie2026llada2,
  title   = {Llada2. 1: Speeding up text diffusion via token editing},
  author  = {Bie, Tiwei and Cao, Maosong and Cao, Xiang and Chen, Bingsen and Chen, Fuyuan and Chen, Kun and Du, Lun and Feng, Daozhuo and Feng, Haibo and Gong, Mingliang and others},
  journal = {arXiv preprint arXiv:2602.08676},
  year    = {2026}
}

@article{lou2023discrete,
  title   = {Discrete diffusion modeling by estimating the ratios of the data distribution},
  author  = {Lou, Aaron and Meng, Chenlin and Ermon, Stefano},
  journal = {arXiv preprint arXiv:2310.16834},
  year    = {2023}
}

@misc{yuan2023scalingrelationshiplearningmathematical,
  title         = {Scaling Relationship on Learning Mathematical Reasoning with Large Language Models},
  author        = {Zheng Yuan and Hongyi Yuan and Chengpeng Li and Guanting Dong and Keming Lu and Chuanqi Tan and Chang Zhou and Jingren Zhou},
  year          = {2023},
  eprint        = {2308.01825},
  archiveprefix = {arXiv},
  primaryclass  = {cs.CL},
  url           = {https://arxiv.org/abs/2308.01825}
}

@article{dong2023raft,
  title   = {Raft: Reward ranked finetuning for generative foundation model alignment},
  author  = {Dong, Hanze and Xiong, Wei and Goyal, Deepanshu and Zhang, Yihan and Chow, Winnie and Pan, Rui and Diao, Shizhe and Zhang, Jipeng and Shum, Kashun and Zhang, Tong},
  journal = {arXiv preprint arXiv:2304.06767},
  year    = {2023}
}

@article{xiong2025minimalist,
  title   = {A minimalist approach to llm reasoning: from rejection sampling to reinforce},
  author  = {Xiong, Wei and Yao, Jiarui and Xu, Yuhui and Pang, Bo and Wang, Lei and Sahoo, Doyen and Li, Junnan and Jiang, Nan and Zhang, Tong and Xiong, Caiming and others},
  journal = {arXiv preprint arXiv:2504.11343},
  year    = {2025}
}

@article{gulcehre2023reinforced,
  title   = {Reinforced self-training (rest) for language modeling},
  author  = {Gulcehre, Caglar and Paine, Tom Le and Srinivasan, Srivatsan and Konyushkova, Ksenia and Weerts, Lotte and Sharma, Abhishek and Siddhant, Aditya and Ahern, Alex and Wang, Miaosen and Gu, Chenjie and others},
  journal = {arXiv preprint arXiv:2308.08998},
  year    = {2023}
}

@article{zelikman2022star,
  title   = {Star: Bootstrapping reasoning with reasoning},
  author  = {Zelikman, Eric and Wu, Yuhuai and Mu, Jesse and Goodman, Noah},
  journal = {Advances in Neural Information Processing Systems},
  volume  = {35},
  pages   = {15476--15488},
  year    = {2022}
}

@article{schulman2017proximal,
  title   = {Proximal policy optimization algorithms},
  author  = {Schulman, John and Wolski, Filip and Dhariwal, Prafulla and Radford, Alec and Klimov, Oleg},
  journal = {arXiv preprint arXiv:1707.06347},
  year    = {2017}
}

@article{shao2024deepseekmath,
  title   = {Deepseekmath: Pushing the limits of mathematical reasoning in open language models},
  author  = {Shao, Zhihong and Wang, Peiyi and Zhu, Qihao and Xu, Runxin and Song, Junxiao and Bi, Xiao and Zhang, Haowei and Zhang, Mingchuan and Li, YK and Wu, Yang and others},
  journal = {arXiv preprint arXiv:2402.03300},
  year    = {2024}
}

@inproceedings{kang2018self,
  title        = {Self-attentive sequential recommendation},
  author       = {Kang, Wang-Cheng and McAuley, Julian},
  booktitle    = {2018 IEEE international conference on data mining (ICDM)},
  pages        = {197--206},
  year         = {2018},
  organization = {IEEE}
}

@inproceedings{guo2026leveraging,
  title     = {Leveraging the Spatial Hierarchy: Coarse-to-fine Trajectory Generation via Cascaded Hybrid Diffusion},
  author    = {Guo, Baoshen and Hong, Zhiqing and Li, Junyi and Wang, Shenhao and Zhao, Jinhua},
  booktitle = {Proceedings of the 32nd ACM SIGKDD Conference on Knowledge Discovery and Data Mining V. 1},
  pages     = {359--370},
  year      = {2026}
}

@article{yuan2024generating,
  title     = {Generating daily activities with need dynamics},
  author    = {Yuan, Yuan and Ding, Jingtao and Wang, Huandong and Jin, Depeng},
  journal   = {ACM Transactions on Intelligent Systems and Technology},
  volume    = {15},
  number    = {2},
  pages     = {1--28},
  year      = {2024},
  publisher = {ACM, New York, NY}
}

\appendix
\section{Limitations}
\label{app:limitations}

While TrajDLM improves trajectory generation in both fidelity and efficiency, several limitations remain. First, TrajDLM operates over discrete road segment tokens and therefore requires map-matched trajectories at both training and inference time, rather than consuming raw GPS traces directly. This introduces an additional preprocessing step and ties downstream performance to the quality of the upstream map-matching algorithm. Continuous space models avoid this requirement altogether, albeit at the cost of weaker topological consistency.

Second, we evaluate TrajDLM using a single backbone family, \texttt{Qwen3-0.6B} adapted via BD3-LM~\cite{arriola2025block,zhou2026dllm}. Alternative LLM backbones, discrete diffusion variants, or model scales may yield different performance trade-offs. For instance, larger models may achieve higher fidelity but at the cost of efficiency, while smaller models may be more efficient but less accurate.

Finally, TrajDLM generates sequences of road segments without explicit temporal modeling. While this is sufficient for the spatial fidelity metrics considered in this work, it limits applications requiring temporally realistic trajectories. Extending block diffusion language models to jointly model spatio-temporal trajectories is another promising direction for future research.

\section{Topology-Constrained Sampling Details and Pseudocode}
\label{app:tcs-pseudocode}

\begin{algorithm}[t]
    \caption{Topology-Constrained Sampling}
    \label{alg:tcs}
    \begin{algorithmic}
        \footnotesize
        \State \textbf{Input:} prompt $x$, model $G_\theta$, blocks $B$, block length $L'$, adjacency penalty $\boldsymbol{P}$, vocabulary mask $\boldsymbol{m}^{\text{voc}}$, origin $r_{\text{org}}$, destination $r_{\text{dest}}$, segment costs $c$, remaining costs $h$ to $r_{\text{dest}}$, guidance weights $w_{\text{dest}}, w_{\text{far}}$, ratio $\rho$, cap $\Delta_{\max}$, temperature $\lambda$
        \vspace{3pt}
        \State $\hat{\tau} \gets \emptyset$;\; $c_{\text{trav}} \gets 0$ \Comment{distance travelled}
        \For{$b = 1$ to $B$}
        \State $\hat{\tau}^b \gets [\texttt{[M]}]^{L'}$ \Comment{initialize block with mask tokens}
        \For{$i = 1$ to $L'$} \Comment{commit one position per step, left to right}
        \State $\boldsymbol{f} \gets G_\theta(x \oplus \hat{\tau}^{<b} \oplus \hat{\tau}^b)_i + \boldsymbol{m}^{\text{voc}}$ \Comment{road + structural tokens only}
        \If{$\hat{r}_{i-1} = \texttt{[TRAJ\_START]}$}
        \State $\boldsymbol{f} \gets -\infty$;\; $f_{r_{\text{org}}} \gets 0$ \Comment{origin anchoring}
        \ElsIf{$\hat{r}_{i-1}$ is a road segment}
        \State $\boldsymbol{f} \gets \boldsymbol{f} + \boldsymbol{P}_{\hat{r}_{i-1}, :}$; only adjacent roads or end tokens allowed \Comment{adjacency}
        \State mask end tokens \Comment{destination reaching}
        \State $w \gets w_{\text{far}}$ \textbf{if} $c_{\text{trav}} > \rho \, h(r_{\text{org}})$ \textbf{else} $w_{\text{dest}}$
        \State $f_r \gets f_r - w \min\big(c(r) + h(r) - h(\hat{r}_{i-1}), \Delta_{\max}\big)$ for all roads $r$ \Comment{dest. guidance}
        \State \textbf{if} some allowed road is not in $\hat{\tau} \oplus \hat{\tau}^b_{1:i-1}$ \textbf{then} mask roads in $\hat{\tau} \oplus \hat{\tau}^b_{1:i-1}$ \Comment{revisit avoidance}
        \Else
        \State mask all road segments \Comment{a road may only follow \texttt{[TRAJ\_START]} or a road}
        \EndIf
        \State $\hat{r}_i \gets \arg\max_{r} \big( f_{r} + \lambda \, g_{r} \big),\; g_{r} \sim \mathrm{Gumbel}(0,1)$;\; commit $\hat{r}_i$ into $\hat{\tau}^b$
        \State $c_{\text{trav}} \gets c_{\text{trav}} + c(\hat{r}_i)$ \textbf{if} $\hat{r}_i$ is a road
        \If{$\hat{r}_i = r_{\text{dest}}$} \Return $\hat{\tau} \oplus \hat{\tau}^b_{1:i}$ \EndIf
        \EndFor
        \State $\hat{\tau} \gets \hat{\tau} \oplus \hat{\tau}^b$
        \EndFor
        \State \Return $\hat{\tau}$
    \end{algorithmic}
\end{algorithm}

This appendix details the topology-constrained sampling (TCS) strategy summarized in Section~\ref{sec:tcd}.

\paragraph{Adjacency Penalty}

Let $\boldsymbol{A} \in \{0,1\}^{|\mathcal{V}| \times |\mathcal{V}|}$ be the road adjacency matrix with $A_{ij} = 1$ iff $(r_i, r_j) \in \mathcal{E}$. We convert $\boldsymbol{A}$ into a log-space penalty matrix
\begin{equation}
    P_{ij} = \begin{cases} 0 & \text{if } A_{ij} = 1, \\ -\infty & \text{otherwise}, \end{cases}
\end{equation}
implemented in practice as a large negative constant for numerical stability. Row $\boldsymbol{P}_{i,:}$ thus encodes which road segments are directly reachable from $r_i$.

\paragraph{Constrained Block Sampling}

During the reverse process, we generate tokens sequentially within each block to preserve local consistency. Let $\boldsymbol{f}_i \in \mathbb{R}^{|\mathcal{V}|}$ denote the logits at position $i \in \{1, \ldots, L'\}$ of the current block. Instead of independent parallel sampling, we adopt a left-to-right decoding scheme so that each prediction conditions on the previously generated segment $\hat{r}_{i-1}$. We apply the adjacency penalty and the destination guidance $\boldsymbol{D}_i$ to the logits and sample via temperature-scaled Gumbel-max:
\begin{equation}
    \hat{r}_i = \operatorname*{arg\,max}_{r \in \mathcal{V}} \left( f_{i,r} + P_{\hat{r}_{i-1}, r} - D_{i,r} + \lambda \, g_{i,r} \right), \quad g_{i,r} \sim \mathrm{Gumbel}(0, 1),
\end{equation}
where $\lambda \geq 0$ is the sampling temperature. This guarantees that invalid transitions receive zero probability, ensuring $(\hat{r}_{i-1}, \hat{r}_i) \in \mathcal{E}$ at every step. A vocabulary mask further restricts generation to road segment tokens and the structural tokens of the trajectory format. After a road segment, only an adjacent road segment may follow.

\paragraph{Left-to-Right Commitment}

At each denoising step, positions are committed strictly from left to right, such that masked positions always form a suffix of the block and each prediction is conditioned on an already committed predecessor. This ensures that every committed segment is adjacent to its predecessor, preserving topological validity by construction. For all experiments, each block of length $L' = 64$ is denoised in 64 steps, committing exactly one position per step.

\paragraph{Origin Anchoring and Destination Reaching}

Given the conditioning origin $r_{\text{org}}$ and destination $r_{\text{dest}}$, the first road segment after the start-of-trajectory token is fixed to $r_{\text{org}}$, and a road segment may only follow the start-of-trajectory token or another road segment, so no trajectory can bypass the origin. End-of-trajectory tokens are masked until $r_{\text{dest}}$ has been generated, and generation stops as soon as it is. Every trajectory therefore satisfies $\hat{r}_1 = r_{\text{org}}$ and $\hat{r}_{|\hat{\tau}|} = r_{\text{dest}}$.

\paragraph{Destination Guidance}

We add a soft, A$^*$-style guidance term that penalises the extra distance a candidate adds over the shortest way to the destination. Let $c(r)$ be the length of segment $r$ in units of the median segment length, and let $h(r)$ be the shortest remaining cost from $r$ to $r_{\text{dest}}$ under $c$ (computed once per trajectory with Dijkstra's algorithm on the reversed road graph). The detour of a candidate $r$ after predecessor $\hat{r}_{i-1}$ is
\begin{equation}
    \delta(r \mid \hat{r}_{i-1}) = \min\big(c(r) + h(r) - h(\hat{r}_{i-1}),\; \Delta_{\max}\big) \geq 0,
\end{equation}
which is zero exactly on shortest paths to the destination. The guidance is $D_{i,r} = w_i\, \delta(r \mid \hat{r}_{i-1})$, with $w_i = w_{\text{dest}}$ until the distance travelled so far exceeds $\rho$ times the shortest origin--destination distance, after which $w_i = w_{\text{far}} \gg w_{\text{dest}}$ pulls long detours back towards the destination. Because $\boldsymbol{D}_i$ is finite, it only re-weights adjacency-valid candidates and never overrides the model entirely.

\paragraph{Revisit Avoidance}

During generation, road segments already visited in the current trajectory are masked to prevent revisits. If all adjacent segments have been visited, the mask is lifted to ensure that generation can continue.

\paragraph{Destination-Aware Termination}

Once all blocks are denoised, we enforce trajectory-level validity by ensuring termination at the destination. We identify the first occurrence of $r_{\text{dest}}$, $i^\star = \min \{\, i : \hat{r}_i = r_{\text{dest}} \,\}$, and truncate the sequence by setting $\hat{r}_i = \texttt{[EOS]}$ for all $i > i^\star$. The resulting sequence $\hat{\tau}$ therefore satisfies $\hat{r}_{|\hat{\tau}|} = r_{\text{dest}}$ and $(\hat{r}_{i-1}, \hat{r}_i) \in \mathcal{E}$ for every intermediate transition.

\paragraph{Pseudocode}

We show the pseudocode for the TCS introduced in Section~\ref{sec:tcd} in Algorithm~\ref{alg:tcs}. Within each block, positions are committed strictly left to right. For the next masked position, the procedure \ding{182} computes logits with $G_\theta$ and restricts the vocabulary, \ding{183} enforces origin anchoring, \ding{184} applies the adjacency penalty $\boldsymbol{P}$ against the committed predecessor and masks end tokens until $r_{\text{dest}}$ is reached, \ding{185} subtracts the destination guidance $\boldsymbol{D}$, \ding{186} masks already visited segments when an unvisited successor exists, and \ding{187} samples and commits the token. Generation stops once $r_{\text{dest}}$ has been generated.

\section{TrajRFT Details and Pseudocode}
\label{app:rft-pseudocode}

\paragraph{Motivation}
The SFT model with TCS generates valid trajectories, but its errors concentrate in long trips, where it commits early to a corridor that diverges from the real route. Across San Francisco, Porto, and Beijing, the longest quarter of trips (at least $46$, $51$, and $39$ road segments, respectively) accounts for $67$--$69\%$ of the total DTW, and the worst $5\%$ of trips for $40$--$42\%$. On these worst trips, the generated route diverges from the real one within the first $2$--$4$ road segments and reaches the destination via a longer path. Sampling from the SFT model at a non-zero temperature, however, reveals that it can already generate far more realistic routes than its greedy decoding. For instance, on the first $100$ San Francisco test trips, picking the best of $K=8$ routes sampled at temperature $1.0$ (i.e., the TrajRFT rollout setting) by DTW against the real route lowers DTW from $13.43$ (greedy decoding; temperature 0.0) to $5.87$. TrajRFT turns this gap between greedy decoding and best-of-$K$ sampling into a training signal via rejection-sampling fine-tuning.

\paragraph{Pseudocode}
Algorithm~\ref{alg:rft} presents the pseudocode for TrajRFT, the rejection-sampling fine-tuning procedure described in Section~\ref{sec:rft}.

\begin{algorithm}[t]
    \caption{Trajectory Rejection-Sampling Fine-Tuning (TrajRFT)}
    \label{alg:rft}
    \begin{algorithmic}
        \footnotesize
        \State \textbf{Input:} SFT model $G_{\theta_0}$, training trajectories $\mathcal{T}$, steps $N$, batch size $|\mathcal{B}|$, samples per trip $K$, margin $\epsilon$, KL weight $\gamma$, scored segments per route $M$
        \vspace{3pt}
        \State $\theta \gets \theta_0$ \Comment{start from SFT; $G_{\theta_0}$ is the frozen KL reference}
        \State freeze the RNE and its projection
        \For{$n = 1$ to $N$}
        \State sample a batch $\mathcal{B} = \{(x_p, \tau_p)\}$ of trips from $\mathcal{T}$, each with probability $\propto |\tau_p|$;\; $\mathcal{S} \gets \emptyset$
        \For{$p = 1$ to $|\mathcal{B}|$}
        \State $\hat{\tau}^{(0)} \gets \mathrm{TCS}(G_\theta, x_p, \lambda{=}0)$ \Comment{greedy route}
        \State $\hat{\tau}^{(k)} \gets \mathrm{TCS}(G_\theta, x_p, \lambda{=}1)$ for $k = 1, \dots, K$ \Comment{sampled routes}
        \State $k^\star \gets \arg\min_{k \geq 1} \mathrm{DTW}(\hat{\tau}^{(k)}, \tau_p)$
        \If{$\mathrm{DTW}(\hat{\tau}^{(k^\star)}, \tau_p) < (1 - \epsilon)\, \mathrm{DTW}(\hat{\tau}^{(0)}, \tau_p)$}
        \State $\mathcal{S} \gets \mathcal{S} \cup \{\hat{\tau}^{(k^\star)}\}$
        \EndIf
        \EndFor
        \State score up to $M$ road segments of every $\hat{\tau} \in \mathcal{S}$ in the state it was sampled from, under the constrained distribution of TCS, with $G_\theta$ and $G_{\theta_0}$
        \State update $\theta$ with an AdamW step on $\mathcal{L}_{\mathrm{RFT}}(\theta)$ \Comment{Eq.~\ref{eq:rft}}
        \EndFor
        \State \Return $G_\theta$
    \end{algorithmic}
\end{algorithm}

\section{Dataset Statistics}
\label{app:datasets}

We train and evaluate TrajDLM on three city-scale trajectory datasets from Beijing, Porto, and San Francisco, following the same preprocessed datasets released by HOSER~\cite{cao2025hoser}. In these datasets, raw GPS trajectories have been map-matched onto road network graphs extracted from OpenStreetMap, converting continuous GPS coordinates into sequences of road segments. We use the same train/validation/test splits provided by HOSER.

For cross-domain evaluation, we additionally use GeoLife~\cite{zheng2010geolife}, which contains trajectories collected in Beijing between 2007 and 2011\footnote{Although~\cite{zheng2010geolife} states that GeoLife was collected up to October 2011, the released dataset contains trajectories with timestamps extending to July 2012.} across diverse transportation modes. We map-match GeoLife trajectories using Fast Map Matching (FMM)~\cite{Yang2018FastMM} and the Beijing road network graph used in HOSER. Map matching filters out a large fraction of trajectories with significant GPS errors or invalid map-matching results. We further filter the dataset by retaining only trajectories with at most 64 road segments, trip distances of at least 1 km, and trip durations between 2 minutes and 2 hours.

Table~\ref{tab:dataset-stats} provides the statistics of all datasets, while Table~\ref{tab:geolife-modes} reports the transportation-mode distribution of the filtered GeoLife dataset.

\begin{table}[htbp]
    \centering
    \scriptsize
    \caption{\textbf{Dataset statistics for all trajectory datasets}.}
    \label{tab:dataset-stats}
    \setlength{\tabcolsep}{6pt}
    \begin{tabular}{lcccc}
        \toprule
        \textbf{Statistics}    & \textbf{Beijing} & \textbf{Porto} & \textbf{San Francisco} & \textbf{GeoLife} \\
        \midrule
        \#roads                & 40,060           & 11,024         & 27,187                 & 40,060           \\
        \#trajectories         & 899,115          & 687,656        & 293,023                & 1,071            \\
        Mean trajectory length & 28.31            & 40.04          & 36.12                  & 28.63            \\
        Max trajectory length  & 60               & 276            & 271                    & 64               \\
        Mean time interval (s) & 28.26            & 12.28          & 15.88                  & 32.41            \\
        Start date             & 01-11-2015       & 01-07-2013     & 17-05-2008             & 14-04-2007       \\
        End date               & 08-11-2015       & 01-07-2014     & 10-06-2008             & 27-07-2012       \\
        \bottomrule
    \end{tabular}
\end{table}

\begin{table}[htbp]
    \centering
    \small
    \caption{\textbf{Transportation-mode distribution of the filtered GeoLife dataset}.}
    \label{tab:geolife-modes}
    \setlength{\tabcolsep}{8pt}
    \begin{tabular}{lc}
        \toprule
        \textbf{Transport mode} & \textbf{\#trajectories} \\
        \midrule
        Unclassified            & 929                     \\
        Bike                    & 69                      \\
        Walk                    & 36                      \\
        Car                     & 24                      \\
        Bus                     & 7                       \\
        Taxi                    & 3                       \\
        Subway / Light rail     & 3                       \\
        \midrule
        Total                   & 1,071                   \\
        \bottomrule
    \end{tabular}
\end{table}

\section{Implementation Details}
\label{app:implementation-details}

We implement TrajDLM with the dLLM library~\cite{zhou2026dllm} and PyTorch Scaled Dot-Product Attention (SDPA), initializing from the checkpoint in Section~\ref{sec:experimental-setup}. SFT uses AdamW with weight decay $0.01$, and TrajRFT reuses the same model and sampler. Table~\ref{tab:implementation-details} lists all training and inference hyperparameters. All training is performed on NVIDIA H100 GPUs.

\begin{table}[t]
    \centering
    \footnotesize
    \caption{\textbf{Training and inference hyperparameters used for TrajDLM.} GeoLife reuses the Beijing model (zero-shot).}
    \label{tab:implementation-details}
    \setlength{\tabcolsep}{3pt}
    \begin{tabular}{lcccc}
        \toprule
        \textbf{Hyperparameter} & \textbf{Beijing} & \textbf{Porto} & \textbf{SF} & \textbf{GeoLife} \\
        \midrule
        \rowcolor{ivory2}
        \multicolumn{5}{l}{\textit{SFT}} \\
        Block length $L'$ & 64 & 64 & 64 & -- \\
        Max length & 256 & 512 & 512 & -- \\
        Learning rate & $1e-4$ & $1e-4$ & $1e-4$ & -- \\
        Warmup ratio & 0.1 & 0.1 & 0.1 & -- \\
        Num. epochs & 3 & 5 & 5 & -- \\
        Batch size & 32 & 16 & 16 & -- \\
        \midrule
        \rowcolor{ivory2}
        \multicolumn{5}{l}{\textit{TrajRFT}} \\
        Steps $N$ & 3,000 & 3,000 & 3,000 & -- \\
        Batch size $|\mathcal{B}|$ & 32 & 32 & 32 & -- \\
        Samples per trip $K$ & 8 & 8 & 8 & -- \\
        Sampling temperature & 1.0 & 1.0 & 1.0 & -- \\
        Selection metric & DTW & DTW & DTW & -- \\
        Margin $\epsilon$ & 0.05 & 0.05 & 0.05 & -- \\
        KL weight $\gamma$ & 0.02 & 0.02 & 0.02 & -- \\
        Learning rate $\eta_{\max}$ & $8e-6$ & $4e-6$ & $4e-6$ & -- \\
        Learning rate $\eta_{\min}$ & $1e-6$ & $1e-6$ & $1e-6$ & -- \\
        LR schedule & cosine & cosine & cosine & -- \\
        Scored segments $M$ & 192 & 192 & 192 & -- \\
        \midrule
        \rowcolor{ivory2}
        \multicolumn{5}{l}{\textit{Inference}} \\
        Max new tokens & 128 & 512 & 256 & 128 \\
        Denoising steps per block & 64 & 64 & 64 & 64 \\
        Max. denoising steps per trajectory & 128 & 512 & 256 & 128 \\
        Temperature & 0.0 & 0.0 & 0.0 & 0.0 \\
        Guidance $w_{\text{dest}}$ & 0.5 & 0.5 & 0.5 & 1.0 \\
        Guidance $w_{\text{far}}$ & 20 & 20 & 20 & 20 \\
        Far ratio $\rho$ & 1.5 & 1.5 & 1.5 & 1.5 \\
        Detour cap $\Delta_{\max}$ & 8 & 8 & 8 & 8 \\
        \bottomrule
    \end{tabular}
\end{table}

\section{Visualizations}
\label{app:visualizations}

\begin{figure}[htbp]
    \centering
    \begin{subfigure}[t]{0.24\linewidth}
        \centering
        \includegraphics[width=\linewidth]{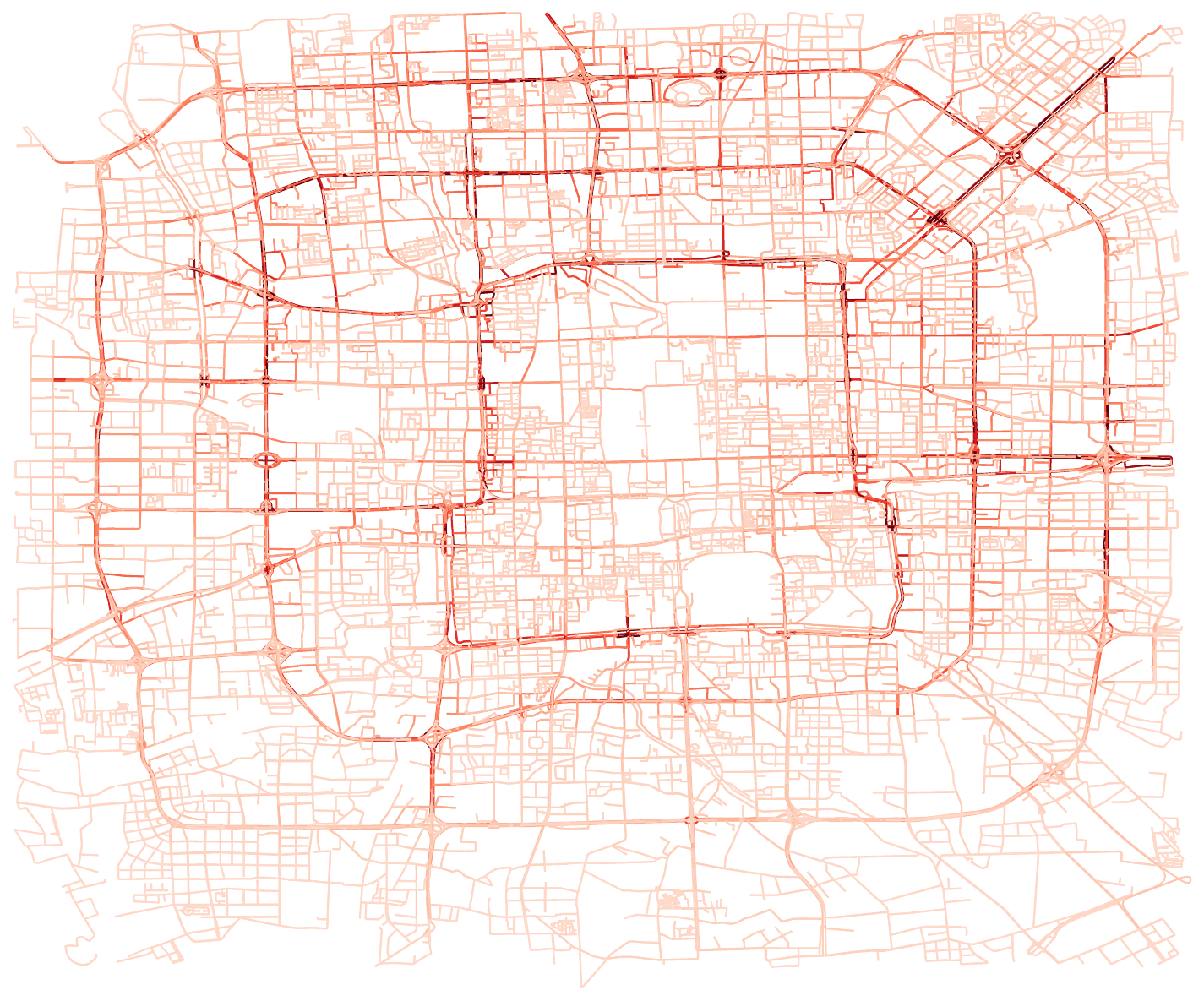}
        \caption{DiffTraj}
    \end{subfigure}
    \hfill
    \begin{subfigure}[t]{0.24\linewidth}
        \centering
        \includegraphics[width=\linewidth]{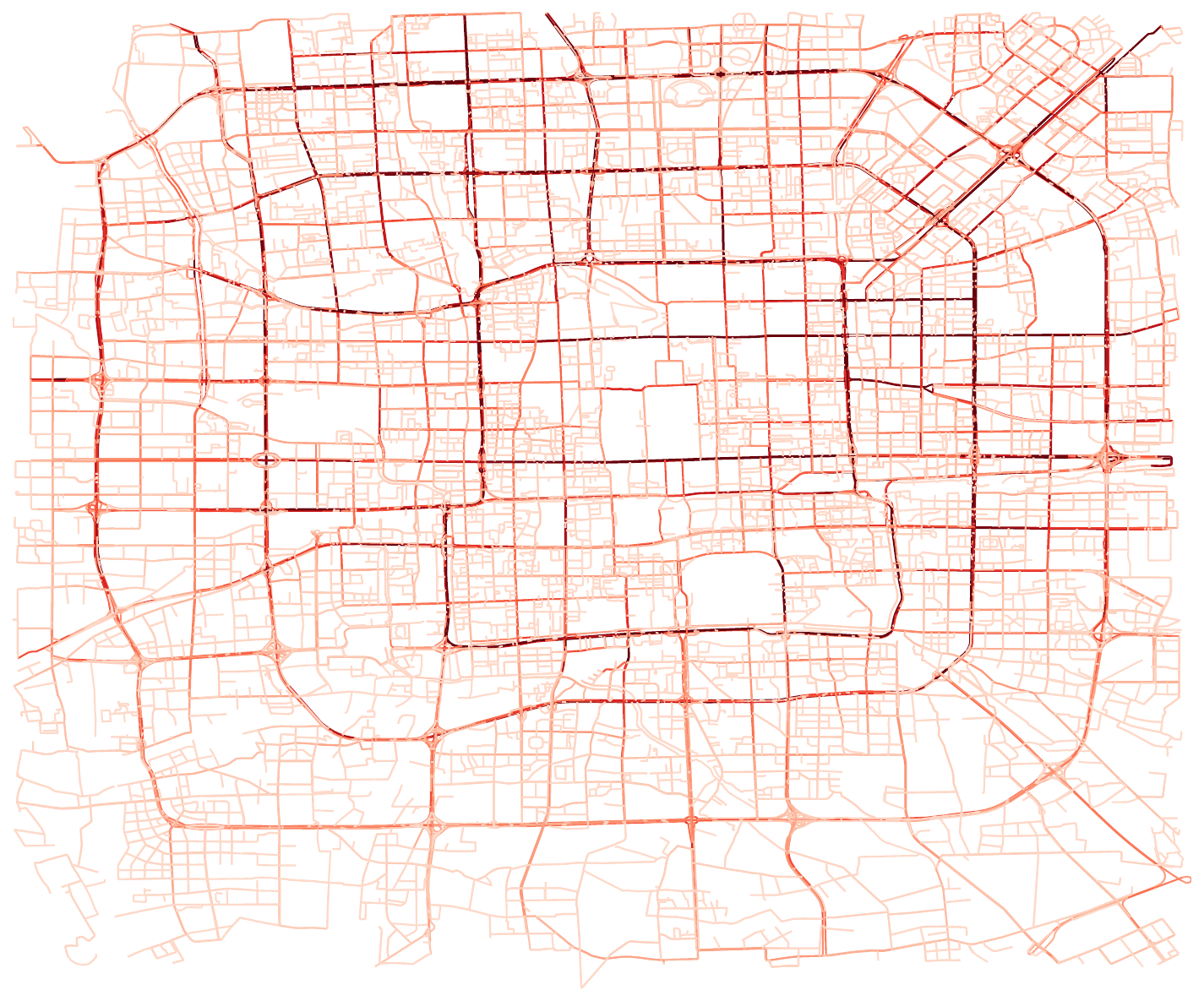}
        \caption{HOSER}
    \end{subfigure}
    \hfill
    \begin{subfigure}[t]{0.24\linewidth}
        \centering
        \includegraphics[width=\linewidth]{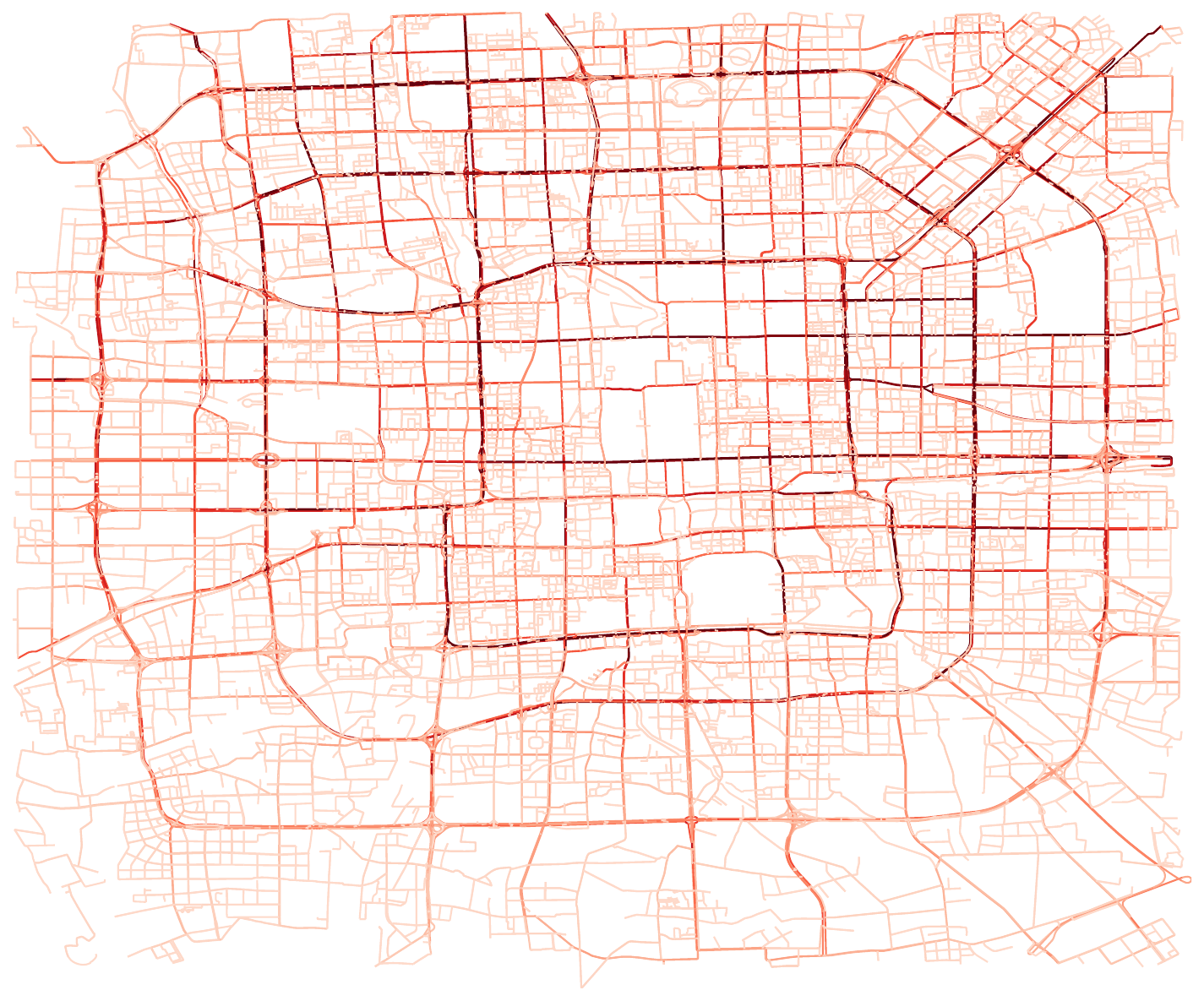}
        \caption{TrajDLM}
    \end{subfigure}
    \hfill
    \begin{subfigure}[t]{0.24\linewidth}
        \centering
        \includegraphics[width=\linewidth]{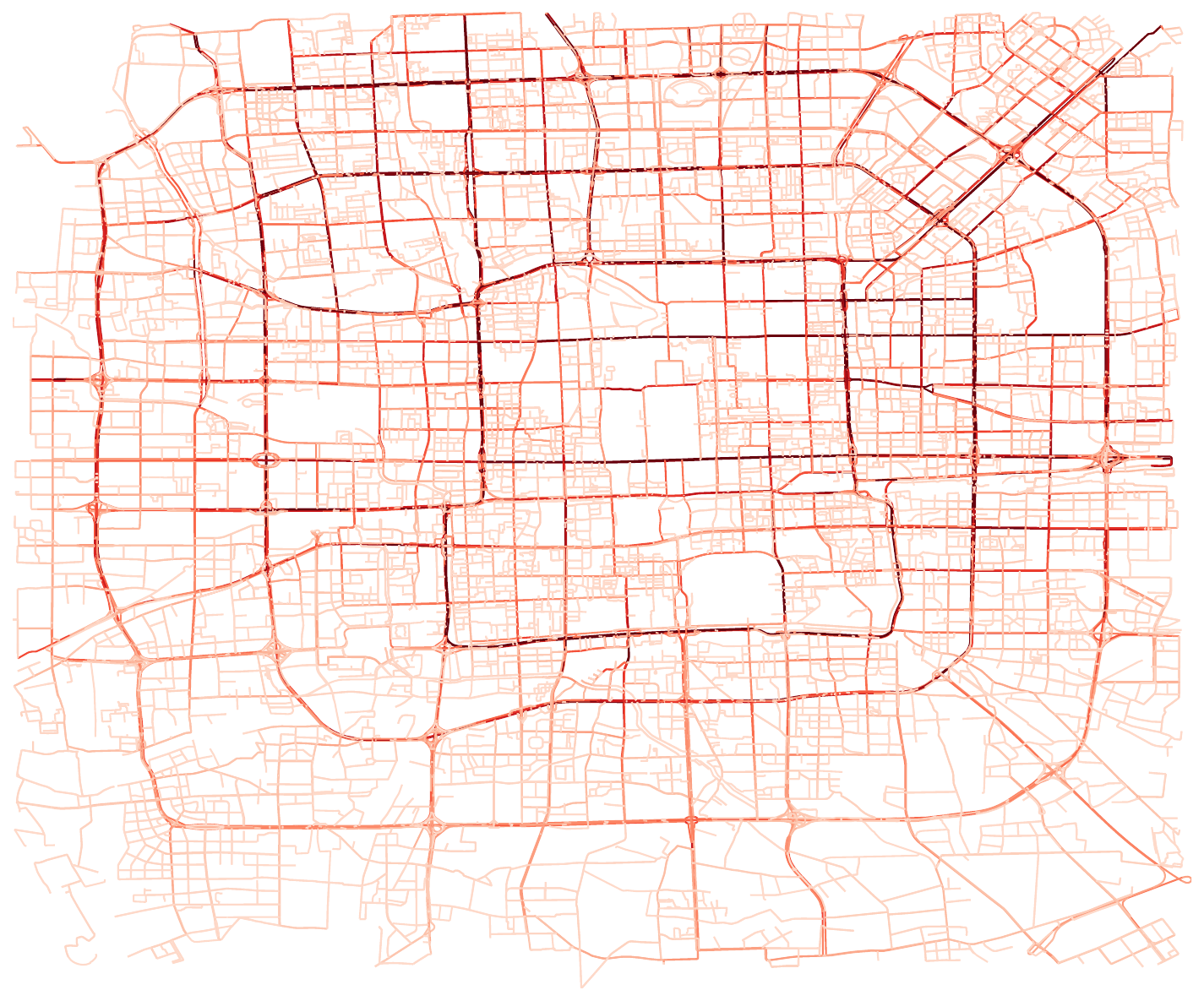}
        \caption{Real}
    \end{subfigure}
    \caption{\textbf{Heatmap visualizations of generated trajectories in Beijing.}}
    \Description{Trajectory density heatmaps on the Beijing road network, for DiffTraj, HOSER, TrajDLM, and real trajectories. Each road segment is colored by trajectory density, with darker red indicating more trajectories.}
    \label{fig:heatmap_beijing}
\end{figure}

\begin{figure}[htbp]
    \centering
    \begin{subfigure}[t]{0.24\linewidth}
        \centering
        \includegraphics[width=\linewidth]{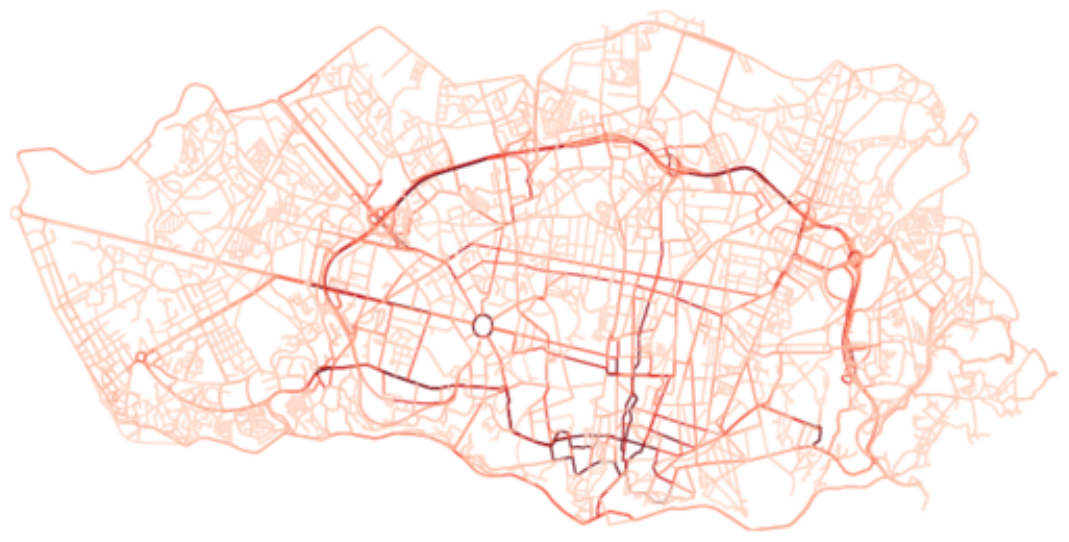}
        \caption{DiffTraj}
    \end{subfigure}
    \hfill
    \begin{subfigure}[t]{0.24\linewidth}
        \centering
        \includegraphics[width=\linewidth]{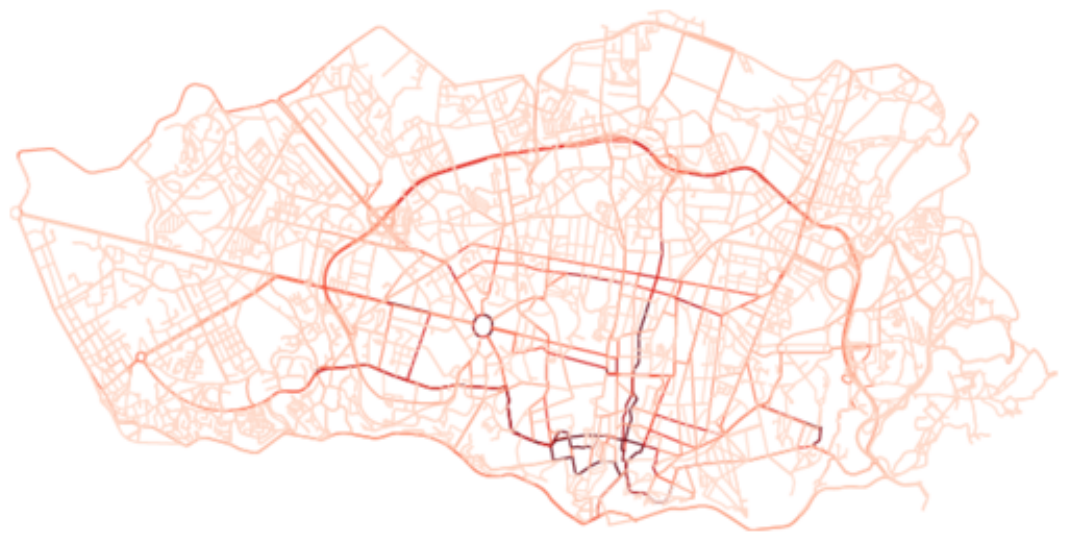}
        \caption{HOSER}
    \end{subfigure}
    \hfill
    \begin{subfigure}[t]{0.24\linewidth}
        \centering
        \includegraphics[width=\linewidth]{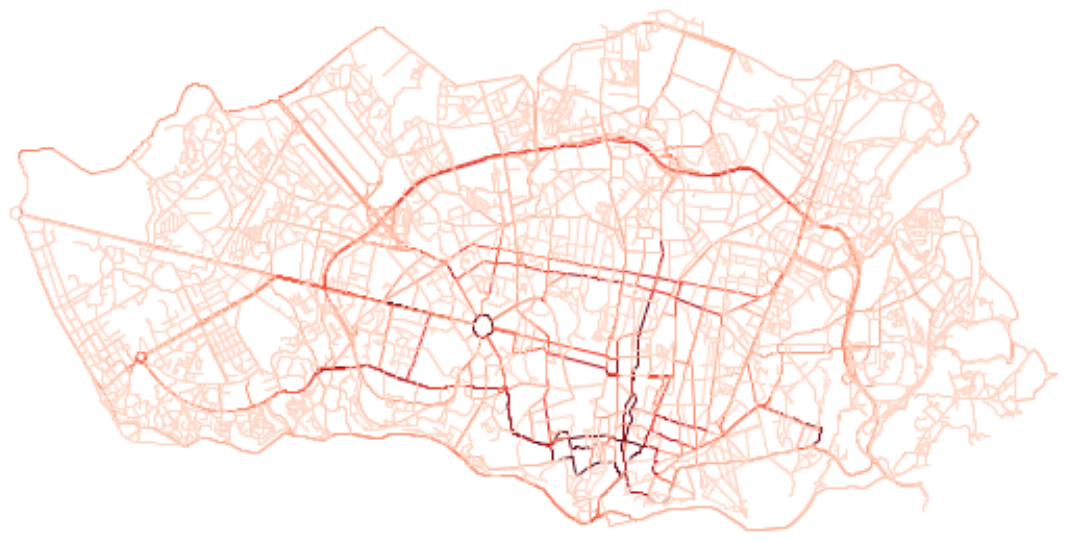}
        \caption{TrajDLM}
    \end{subfigure}
    \hfill
    \begin{subfigure}[t]{0.24\linewidth}
        \centering
        \includegraphics[width=\linewidth]{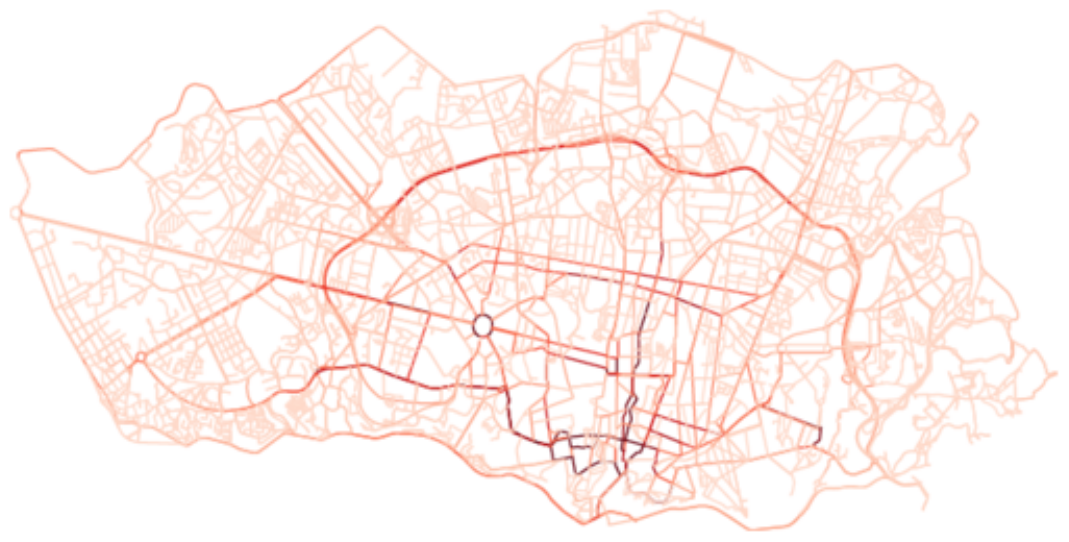}
        \caption{Real}
    \end{subfigure}
    \caption{\textbf{Heatmap visualizations of generated trajectories in Porto.}}
    \Description{Trajectory density heatmaps on the Porto road network, for DiffTraj, HOSER, TrajDLM, and real trajectories. Each road segment is colored by trajectory density, with darker red indicating more trajectories.}
    \label{fig:heatmap_porto}
\end{figure}

We present trajectory heatmaps for Beijing and Porto, comparing ground-truth trajectories with those generated by DiffTraj, HOSER, and TrajDLM in Figs.~\ref{fig:heatmap_beijing}--\ref{fig:heatmap_porto}. Across all, TrajDLM produces heatmaps that closely align with the spatial distribution of the real data, capturing both high-density regions and overall mobility patterns. In contrast, models such as DiffTraj, which generate trajectories in continuous coordinate space without explicit road network constraints, tend to produce trajectories in regions that are infrequently visited in the real data. This discrepancy reflects their inability to fully respect the underlying road network topology. Overall, these visualizations qualitatively confirm that TrajDLM generates trajectories that are both spatially coherent and consistent with real-world mobility patterns across diverse urban environments.

\clearpage %

\end{document}